\documentclass{article}

\usepackage{arxiv}
\usepackage[utf8]{inputenc}
\usepackage[T1]{fontenc}

\usepackage{amsmath}
\usepackage{amsfonts}
\usepackage{booktabs}
\usepackage{graphicx}
\usepackage{longtable}
\usepackage{array}
\usepackage{calc}
\usepackage{fancyvrb}
\usepackage{microtype}
\usepackage{natbib}
\usepackage{textcomp}
\usepackage{hyperref}
\usepackage{xurl}
\usepackage{doi}
\makeatletter
\newcommand{\needspace}[1]{\par\penalty-100\begingroup
  \setlength{\dimen@}{#1}\dimen@ii\pagegoal\advance\dimen@ii-\pagetotal
  \ifdim\dimen@>\dimen@ii\ifdim\dimen@ii>\z@\vfil\fi\break\fi\endgroup}
\makeatother

\providecommand{\tightlist}{%
  \setlength{\itemsep}{0pt}\setlength{\parskip}{0pt}}

\title{Learning by Doing: Autonomously Compiling and Curating Agent Skills from Verified Traces}
\author{Daniel Chalef \quad Peter Evans \quad Pavlo Paliychuk \\ \textbf{Mehul Patel \quad Preston Rasmussen \quad Jack Ryan} \\[4pt] Zep AI}
\date{}

\hypersetup{
  pdftitle={Learning by Doing: Autonomously Compiling and Curating Agent Skills from Verified Traces},
  pdfauthor={Daniel Chalef, Peter Evans, Pavlo Paliychuk, Mehul Patel, Preston Rasmussen, Jack Ryan},
}

\begin{document}
\maketitle

\begin{abstract}
Memory systems for production LLM agents mostly store semantic memory (facts about users and domains) or episodic memory (earlier interactions). Research systems such as Trace2Skill, WikiSkill, and SkillRevise learn procedures from agent runs, package them as Skills, and evaluate them on benchmarks. We describe Zep Agent Skills, a production service that gives deployed agents this procedural memory. The service uses only verified attempts as evidence, cites the supporting events for each field of a Skill, and governs the admission and distribution of each Skill version. The service records each task attempt as a Trace with ordered events, an outcome, and a verification strength. A compiler with separate Reflector and Curator roles turns verified Traces into versioned Skills. Scans and admission gates control which candidates become active, and verified use outcomes change the search rank of each Skill. The Agent Skills format is an open standard that Anthropic originated and donated to the Agentic AI Foundation, so any host of the format can load a compiled Skill. Agents retrieve Skills through lexical search with progressive disclosure, over an API or MCP. We evaluate Zep Agent Skills on \ensuremath{\tau}\ensuremath{^3}-bench banking, AppWorld, and EnterpriseOps-Gym. Skills compiled from the Traces of a teacher model increase the pass rate of three weaker executor models on the same tasks, at every reasoning effort, by 4.5 to 18.8 points. Most of the gain is on the tasks that the teacher solved. On ALFWorld, Skills compiled from the runs of a small model give a gain similar to or larger than the gain that prior work reports for continuous fine-tuning, with no training.
\end{abstract}

\needspace{10\baselineskip}
\section{Introduction}\label{introduction}

Engineers can write a prompt or a Skill for a stable, known process. A deployed agent also meets variations and exceptions of a process, changes to a process over time, and processes that are new. A support agent gets a ticket type that its instructions do not cover, and an operations agent must follow a changed approval rule. Each run of a large language model (LLM) agent starts from the same weights and the same prompt, so a successful run does not change the next run. The agent must solve the same problem again, and it can make the same errors again.

Two common answers are fine-tuning and retrieval of earlier transcripts. Fine-tuning on successful runs changes the weights, but it needs training rounds, it is tied to one model, and it is difficult to inspect or to undo. Each change to a process needs a new training round. Retrieval of earlier transcripts (episodic memory) is cheap, but a transcript is long, it carries the values of one instance, and it can carry data that a later user must not see. Procedural memory is a third answer (Section 3.1). Research systems such as Voyager \citep{voyager}, Agent Workflow Memory \citep{awm}, ExpeL \citep{expel}, and ACE \citep{ace} show that procedures learned from experience can increase task success.

The Agent Skills format \citep{agentskills} gives agents a portable package for procedures: a folder with a \texttt{SKILL.md} file, frontmatter with a name and a description, and optional files. Agent harnesses such as Claude Code \citep{claudeskills}, Claude Cowork \citep{cowork}, and OpenAI Codex \citep{codexskills} load a Skill with progressive disclosure: the agent first sees the name and the description, and it reads the body only when the Skill applies. Anthropic published the format as an open standard in December 2025 and donated it to the Agentic AI Foundation (AAIF) of the Linux Foundation in October 2026 \citep{aaif}. The AAIF also holds the Model Context Protocol (MCP) \citep{mcp}. The format defines how a Skill is packaged and loaded. The format does not define how a Skill is learned from runs, verified, versioned, or governed, and in most deployments a human writes each Skill.

A production service that learns Skills has requirements in addition to those of a research system. The service must learn only from verified attempts. Each claim in a Skill must trace to its evidence. A new version must not silently replace a version that works. A Skill must not carry the instance data of one user or organization to another. Deletion of a source must remove its effect. The service must work with any agent framework and any model, and it must accept Traces from existing trace stores such as Braintrust and LangSmith.

We evaluate the service with a teacher-executor protocol. A strong model solves each task, Zep compiles Skills from its verified Traces, and weaker models then do the same tasks with and without the Skills. This protocol models a production pattern: a strong model solves a task one time, and a cheaper model then does it again. Skills increase the pass rate of each executor at each reasoning effort, mostly on the tasks that the teacher solved. The main benchmarks do not test transfer to new tasks.

The contributions of this paper are:

\begin{enumerate}
\def\labelenumi{\arabic{enumi}.}
\tightlist
\item
  A resource model for procedural memory: namespace, Trace, Skill, Skill version, and Skill use (Sections 3 and 4).
\item
  A compiler that turns verified Traces into structured Skill contracts with event citations for each field, and that makes local edits to existing Skills (Section 5.3).
\item
  An admission and lifecycle model: scans, gates, maturity tiers, and a search weight from verified use (Section 5.4).
\item
  A retrieval and delivery path: lexical search, progressive disclosure, and MCP delivery of packages in the Agent Skills format (Section 5.5).
\item
  A teacher-executor evaluation protocol with paired arms and leakage controls, and results on four benchmarks (Sections 6 to 8).
\end{enumerate}

Section 2 reviews related work. Sections 3 to 5 describe the problem and the design. Sections 6 and 7 give the evaluation, Section 8 gives the analysis, and Section 9 gives the limits and the future work.

\needspace{10\baselineskip}
\section{Related work}\label{related-work}

\textbf{Memory for LLM agents.} CoALA \citep{coala} divides agent memory into working, semantic, episodic, and procedural memory. Semantic memory systems store facts about users and domains (MemGPT \citep{memgpt}), in some systems in a knowledge graph with validity times (Graphiti \citep{zep}). Zep \citep{zep} offers a context layer: a temporal knowledge graph that combines structured and unstructured enterprise data, with access control, versioning, and retention management. Semantic memory is one application of this layer. Episodic memory systems retrieve earlier attempts or reflections on them (Reflexion \citep{reflexion}). Zep now also offers Zep Agent Skills, which supplies procedural memory and does not replace the other types.

\textbf{Learning procedures from experience.} Voyager \citep{voyager} stores verified programs as a skill library and improves them with execution feedback. ExpeL \citep{expel} extracts insights from successful and failed attempts. Agent Workflow Memory \citep{awm} induces reusable workflows from trajectories and abstracts the values of each instance. MemP \citep{memp} studies how procedural memory is built, retrieved, and updated. LEGOMem \citep{legomem} places procedural memory in multi-agent systems. Skill-Pro \citep{skillpro} learns reusable skills from interaction experience without weight updates. AutoSkill \citep{autoskill} consolidates interaction experience into skills over many sessions, and Trace2Skill \citep{trace2skill} consolidates many execution trajectories into one skill directory.

\textbf{Curation and revision of skill libraries.} A survey of dynamic skill libraries \citep{dynamicskills} describes a lifecycle of evidence acquisition, proposal, verification and admission, storage, retrieval, maintenance, and governance. It also reports weak verifiers, and retrieval over a flat list of skills that degrades as libraries grow. ACE \citep{ace} separates a generator, a reflector, and a curator that make incremental edits to a context playbook. Anything2Skill \citep{anything2skill} compiles external knowledge, such as manuals, logs, and trajectories, into structured skill contracts in a versioned skill bank. SkillRevise \citep{skillrevise} revises an initial skill from execution traces and keeps the version with the best measured utility. SkillLift \citep{skilllift} learns rubrics from sparse evaluation results to reduce the rollouts that skill revision needs. WikiSkill \citep{wikiskill} and Procedural Graphs \citep{proceduralgraphs} consolidate knowledge and procedure structure across many runs, and both keep a record of rejected proposals for later proposals. Zep Agent Skills follows the lifecycle of the survey. It uses a separate Reflector and Curator, as in ACE, and adds per-field citations, transactional admission, and governance. Unlike Voyager, SkillRevise, and SkillLift, it does not execute candidate Skills (Section 5.4).

\textbf{Portable skill packages.} The Agent Skills format \citep{agentskills} and the draft MCP Skills extension SEP-2640 \citep{sep2640} define how an agent harness discovers and loads a Skill. Zep Agent Skills emits packages in this format.

\textbf{Continuous fine-tuning.} Methods such as Retrospective Self-Distillation (RSD) \citep{rsd} train the weights of a model on its own runs in several rounds. Section 7.6 compares the gain of RSD with the gain of Skills on ALFWorld.

\textbf{Benchmarks.} \ensuremath{\tau}-bench \citep{taubench}, \ensuremath{\tau}\ensuremath{^2}-bench \citep{tau2}, and the \ensuremath{\tau}\ensuremath{^3} banking knowledge domain \citep{tauknowledge}, AppWorld \citep{appworld}, EnterpriseOps-Gym \citep{eog}, ALFWorld \citep{alfworld}, and SkillsBench \citep{skillsbench} measure agents in tool-use environments. SkillsBench measures the effect of authored and self-generated Skills.

\needspace{10\baselineskip}
\section{Problem setting}\label{problem-setting}

\needspace{10\baselineskip}
\subsection{Setting and terms}\label{setting-and-terms}

We consider a deployed agent that performs tasks over time. Each attempt produces observable events and an outcome, and an external system can verify some outcomes. Zep Agent Skills supplies procedural memory \citep{coala}: when to use a procedure, its steps, and when to stop.

Zep Agent Skills does not train model weights, execute Skills or application code, or store hidden model reasoning. A Trace contains only observable events, and self-reflection alone does not verify an outcome. The service turns verified task evidence into cited, versioned procedures. It admits and retrieves these procedures within the access and applicability boundaries of its namespace.

Table 1 defines the terms of this paper. ``Zep Agent Skills'' names the product, and ``the Agent Skills format'' names the open package format.

\needspace{8\baselineskip}
\textbf{Table 1.} Terms.

\begin{longtable}[]{@{}
  >{\raggedright\arraybackslash}p{\dimexpr 0.2149\linewidth - 0.4298\tabcolsep\relax}
  >{\raggedright\arraybackslash}p{\dimexpr 0.7851\linewidth - 1.5702\tabcolsep\relax}@{}}
\toprule\noalign{}
\begin{minipage}[b]{\linewidth}\raggedright
Term
\end{minipage} & \begin{minipage}[b]{\linewidth}\raggedright
Definition
\end{minipage} \\
\midrule\noalign{}
\endhead
\bottomrule\noalign{}
\endlastfoot
Zep Agent Skills & The Zep product that compiles, curates, and distributes Skills. \\
Agent Skills format & The open Skill package format, with a \texttt{SKILL.md} file, frontmatter, and optional files. \\
Namespace & The logical boundary that owns the Traces and Skills of one deployed agent. \\
Trace & A bounded task attempt with an objective, task family, ordered events, outcome, and verification strength. \\
Event & A typed Trace record that provides evidence about an attempt. \\
Outcome & The result that the caller reports: success, partial success, failure, abandonment, or unknown. \\
Verification strength & The level of evidence for an outcome: none, self-reported by the agent, or independent (from a tool, an authorized human, or an external system). Independent levels are strong. \\
Task family & A group of attempts for the same kind of task, declared by the caller or inferred by Zep. \\
Attempt family & An initial Trace and its retries. Compilation counts attempt families, not Traces. \\
Trace summary & An immutable, task-specific summary of one Trace, with event citations. \\
Skill & An invocable procedure for a task family, with use conditions, steps, stop conditions, outputs, and evidence. \\
Skill version & An immutable Skill revision in a version history. \\
Source & A Trace that a compilation run reads as evidence. The cite-or-retain rule decides when a source is consumed and no longer offered to later runs. \\
Proposal & A change that the Reflector writes for the Curator to review. \\
Candidate & A rendered Skill version that waits for admission. Search does not return a candidate. \\
Admission & The decision that makes a candidate the current, retrievable Skill version. \\
Skill use & A record of an agent's selection or use of a Skill in a Trace, with optional outcome feedback. \\
Security domain & The set of namespaces, users, and data that may use the same Skill, for example one organization or one application. \\
\end{longtable}

\needspace{10\baselineskip}
\subsection{Threat model}\label{threat-model}

A production agent acts on what it retrieves. When the retrieved content is a procedure, the agent follows it as instructions, often with tools that change data, send messages, or move money. Memory-poisoning attacks show this risk for agent memory. AgentPoison \citep{agentpoison} puts malicious records with a trigger into the memory or knowledge base of an agent. A later query that contains the trigger retrieves the records, and the records steer the agent to the action of the attacker. MINJA \citep{minja} gets the same result without access to the memory store: an attacker who only sends queries to the agent makes the agent write the malicious records itself. Skills are a direct path for this type of attack. Skill-Inject \citep{skillinject} reports attack success rates of up to 80\% against frontier models through instructions in skill files, and a study of 98,380 public skills found 157 malicious skills \citep{maliciousskills}.

A service that learns Skills from Traces has the same exposure. A Trace contains tool output and user text that an attacker can control, and the service can admit a Skill without a human. We consider five threats:

\begin{enumerate}
\def\labelenumi{\arabic{enumi}.}
\tightlist
\item
  \textbf{Instance-data leakage.} A Skill can copy a name, identifier, credential, or private URL from a Trace.
\item
  \textbf{Instruction injection.} Tool output or user text in a Trace can contain instructions that a compiler model keeps in a Skill.
\item
  \textbf{Unverified or poisoned outcomes.} An agent can report success for a failed run, or a caller can assert verification without authority. An attacker can then make a harmful procedure look successful.
\item
  \textbf{Cross-boundary flow.} Evidence from one namespace or organization can enter a Skill that another principal reads.
\item
  \textbf{Unsafe procedures.} A Skill can direct an agent to change data, send messages, or move money.
\end{enumerate}

Section 5.6 maps each threat to its controls.

\needspace{10\baselineskip}
\section{System overview}\label{system-overview}

Figure 1 shows the five stages of the lifecycle:

\begin{enumerate}
\def\labelenumi{\arabic{enumi}.}
\tightlist
\item
  \textbf{Capture.} The agent harness records an attempt as a Trace with ordered events, an outcome, and verification strength. Zep can also import Traces.
\item
  \textbf{Compile.} The compiler summarizes closed Traces and uses verified attempts to propose new Skills or changes to existing Skills.
\item
  \textbf{Curate.} The Curator checks proposals against evidence and current Skills. Trusted code validates the result.
\item
  \textbf{Admit.} Zep renders and scans each candidate, then applies the admission policy. A human can review candidates when the namespace requires review.
\item
  \textbf{Serve.} The agent searches active Skills, reads the selected procedures, and records their use with the task outcome.
\end{enumerate}

Each recorded use with a strongly verified outcome changes the search weight and the trust value of the Skill (Section 5.4). Each new Trace is evidence for the next compilation. Capture does not immediately create an active Skill. The evidence passes through compilation and admission before an agent can retrieve the procedure.

\begin{figure}
\centering
\includegraphics[width=1\linewidth,height=\textheight,keepaspectratio]{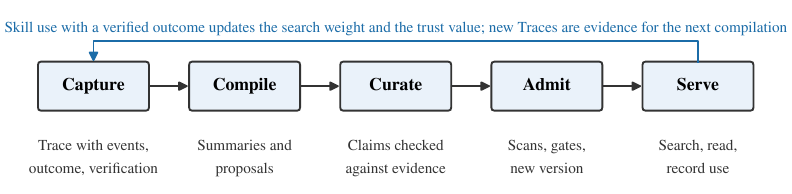}
\caption{Zep Agent Skills lifecycle. A Skill use with a verified outcome updates the search weight and the trust value of the Skill. Each new Trace is evidence for the next compilation.}
\end{figure}

Each namespace owns its Traces and Skills, and each Skill has immutable versions with evidence links. The API exposes capture, retrieval, review, and version management. A human reviewer can inspect evidence and restore an earlier version, and a restore creates a new immutable version. Imports from Braintrust and LangSmith use the same lifecycle as captured Traces and do not create Skills directly. The MCP endpoint serves Claude Code through SEP-2640 and OpenAI Codex through its MCP resource interface.

\needspace{10\baselineskip}
\section{Design of Zep Agent Skills}\label{design-of-zep-agent-skills}

Several stages of the lifecycle use LLMs. Each LLM output is a claim that deterministic code checks before it has an effect. We call this code \emph{trusted code}.

\needspace{10\baselineskip}
\subsection{Traces and verification}\label{traces-and-verification}

The agent harness records ordered events while a task runs, for example model turns, tool calls, and tool results. At the end of the task, the harness closes the Trace with the outcome that it reports.

An agent can report success for a task that failed, so Zep stores the reported outcome separately from its verification. A \emph{verifier} is the party that checks an outcome: the agent itself, a tool, a human, or an external system such as a test suite. Zep sets the verification strength from the identity of the caller that submits the verification, and it ignores a strength that the caller only states. An agent can submit only self-reported verification. A tool or an external system must first be registered as a verifier of the namespace, with the strengths that it may assert. A human must have the permission to verify outcomes. Traces that Zep imports from an LLM observability or evaluation platform, such as Braintrust or LangSmith, can carry the scores of that platform. The import maps a score to an outcome with a threshold that a human sets. When a human registers the import as a verifier, the score is external verification, and it is strong. Zep authenticates the origin of a verification result, but it does not prove that the result is true. When two verification records of a Trace conflict, Zep does not admit Skills from the Trace automatically.

A retry of the task inherits the attempt family of its parent Trace, and compilation counts attempt families. A failure and its successful retry therefore count once as support.

After a Trace closes, an LLM writes a Trace summary. The summary records the objective, the outcome, the approach, the effective and ineffective steps, the errors, the recoveries, and the tools used. A summary is specific to one task and can keep the instance values that explain the attempt. Each claim of the summary cites the Trace events that support it, and trusted code removes each claim that it cannot match to an event. Like the insights of ExpeL \citep{expel}, summaries condense experience. The compiler uses summaries to select relevant attempts, and then it checks each supporting claim against the events. A summary is thus never evidence by itself.

The task family sets the scope of compilation and retrieval: the compiler learns Skills for one family at a time, and search returns the Skills of the requested family. A caller can declare the family. Otherwise, Zep infers the family from the objective, the summary, and the names of the tools and operations of the Trace. The inference input has no tool arguments, tool results, identifiers, or URLs, so instance data cannot decide the family. Zep assigns a known family only when a calibrated score meets a threshold. Otherwise, the Trace stays unassigned and does not enter compilation. When enough distinct attempt families stay unassigned, Zep can propose a new family.

\needspace{10\baselineskip}
\subsection{Skill representation}\label{skill-representation}

Zep stores each Skill version as a structured contract. The contract separates applicability from procedure:

\begin{itemize}
\tightlist
\item
  \textbf{Applicability fields} state when to use the Skill, when not to use it, its preconditions, and the tools and environments where it applies.
\item
  \textbf{Procedure fields} give the ordered steps, the decisions, the expected observations, the outputs, the completion criteria, the stop conditions, and the failure modes with their recovery.
\end{itemize}

The fields follow the activation, execution, and termination conditions of Skill-Pro \citep{skillpro} and the structured contracts of Anything2Skill \citep{anything2skill}. Typed fields let the compiler check and edit one part of a Skill at a time. A field can be empty. For example, the evidence for a new Skill can show no failure mode.

Zep renders each version from the contract into two Markdown forms:

\begin{itemize}
\tightlist
\item
  The \textbf{full form} is for review. It marks the evidence for each cited field.
\item
  The \textbf{agent view} is a \texttt{SKILL.md} package in the Agent Skills format. It has no evidence markers, and it leaves out the sections of empty fields. Agents receive the agent view by default.
\end{itemize}

A human reviewer thus reads the evidence, and the agent reads a procedure without citation markers. Figure 2 shows a compiled Skill in the agent view.

\begin{figure}
\centering
\includegraphics[width=1\linewidth,height=\textheight,keepaspectratio]{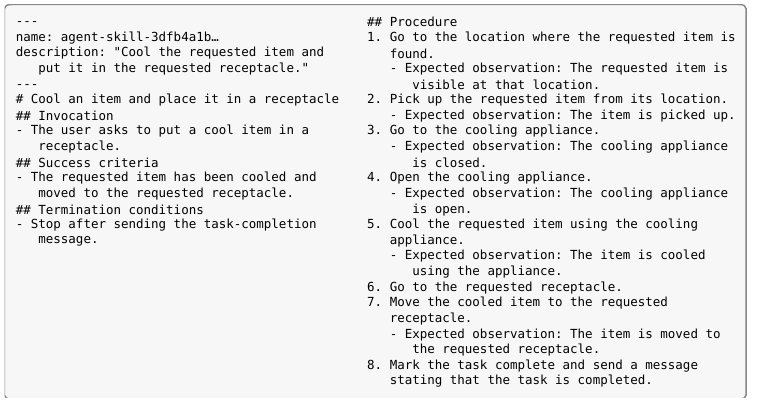}
\caption{Excerpt of a compiled Skill in the agent view, from the ALFWorld experiment. The figure omits some sections, for example the outputs, the tools, and the environments.}
\end{figure}

A simpler design would store each Skill only as a \texttt{SKILL.md} file. A Markdown file does not carry access control, links to evidence, version history, or use outcomes, and a search cannot filter on its fields. Zep therefore stores the contract and renders the \texttt{SKILL.md} file from it.

\needspace{10\baselineskip}
\subsection{Compilation}\label{compilation}

The compiler turns verified Traces into proposed changes to the Skills of one task family. Figure 3 shows the stages. Two LLM roles do the work. The \textbf{Reflector} writes proposals, and the \textbf{Curator} checks the proposals against the evidence and the current Skills. Trusted code checks the output of each role. One model call could write and check a Skill. We separate the roles because claim checking is a narrower task than procedure generation, and separate verdicts give an audit record. The two roles can also use different models or efforts. ACE \citep{ace} uses related generator, reflector, and curator roles.

\begin{figure}
\centering
\includegraphics[width=1\linewidth,height=\textheight,keepaspectratio]{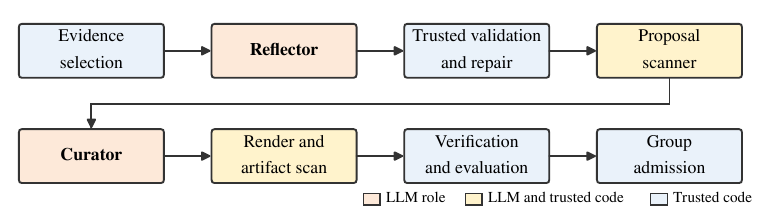}
\caption{Compiler stages. The Reflector and the Curator are LLM roles. Each scanner uses deterministic checks and a model. Trusted code does the other stages.}
\end{figure}

\textbf{Evidence selection.} A compilation run reads the evidence of one namespace and task family. Compilation runs on a schedule, when a Trace closes, or on request. The evidence has four parts:

\begin{itemize}
\tightlist
\item
  the current Skills of the family;
\item
  stale Skills, which are earlier Skills that new evidence contradicts;
\item
  the Trace summaries and their events, from Traces that earlier runs did not consume (see ``Cite-or-retain rule'' below);
\item
  the outcome history of earlier proposals (see ``Outcome history'' below).
\end{itemize}

Only successful attempt families with strong verification can support a new Skill. The selector removes duplicate retries and keeps failed attempts as contrast, as ExpeL \citep{expel} does. Each part has a size limit, and the selector shortens long tool results. It keeps the events that a summary cites, the decisions, and the verification events. Each Trace that a run reads as evidence is a \emph{source} of the run. The run records each source that it read and each source that it left out. A human reviewer can then tell missing evidence from evidence that did not support a proposal.

\textbf{Reflector.} The Reflector can propose five operations: add a Skill, patch a part of a Skill, merge related Skills, report a conflict between new evidence and a current Skill, or make no change. Each proposed claim cites its supporting events. A patch changes only the fields that new evidence supports and keeps the rest of the Skill. Local patches follow ACE \citep{ace}, which reports that full rewrites can lose details. The Reflector replaces instance values, such as names and identifiers, with placeholders.

\textbf{Validation.} Trusted code validates the structure of each proposal and matches each citation to a stored event that the namespace can read. It can repair a citation when the correct event is certain. For example, an event identifier in the text of a claim moves into the citation of the claim. A repair does not change a claim or add evidence, so a repair cannot make an unsupported claim acceptable. Trusted code records each repair and checks support again after it.

A new Skill must apply beyond the values of one instance. When one attempt family supports a Skill, the checks are stricter. The procedure must not keep identifiers from the source events, and its applicability must stay within the tools and environments of the evidence. A failure mode can cite failed attempts. Failed attempts cannot support a procedure step.

\textbf{Scanners.} A scanner uses deterministic checks and a model. It checks content for instance data, such as names, identifiers, credentials, and private URLs, and for instructions that come from tool output. The proposal scanner checks each proposal before the Curator reads it. The artifact scanner checks each rendered candidate before admission. A scanner rejection stops the proposal or the candidate. The summary model, the Reflector, the Curator, and the scanners read Trace content, summaries, and proposals as data, and they do not follow instructions in that content. Model providers give the same rule for tool output \citep{modelspec, injection}. Some events carry content from a restricted resource, for example a document that only some users or agents of the namespace may read. Zep stores such an event as a reference with no content, and the compiler does not read it. No Skill can contain its content. An application that must learn from such evidence uses a separate namespace.

\textbf{Curator.} The Curator gives a verdict of supported or unsupported for each claim, from its citations and the current Skills. Then it decides each proposal. It can add or patch a Skill, merge Skills, replace a Skill with a new Skill, record a conflict, reject the proposal, or make no change. A recorded conflict removes the affected Skill from search until the conflict is resolved. An unsupported claim rejects only its proposed change. The Curator compares each candidate with the current Skills, so that a weaker rewrite does not replace a Skill that works. It prefers a patch of a compatible provisional Skill (Section 5.4) to a second, similar Skill. The Curator returns the related changes of one run as a \emph{result group}, which admission applies together.

\textbf{Cite-or-retain rule.} After a run, the compiler can \emph{consume} a source, and a consumed source is not offered to later runs. The cite-or-retain rule consumes failed sources, unverified sources, and verified sources that an admitted change cites. The rule retains a successful source that no admitted change cites, so that a later run can use it. When the source stays unused, the rule consumes it. Later runs can thus revisit evidence, and unused evidence does not stay in the pool indefinitely.

\textbf{Outcome history.} The compiler records the final outcome of each proposal for each task family: admission, scanner rejection, Curator rejection, human rejection, evaluation failure, or a failed model call. WikiSkill \citep{wikiskill} and Procedural Graphs \citep{proceduralgraphs} report that a record of rejected proposals improves later proposals. The compiler gives the Reflector a short history as untrusted context. The history can help the Reflector avoid a change that was rejected, but it cannot support a proposal. The history excludes deleted sources.

\needspace{10\baselineskip}
\subsection{Admission, maturity, and trust}\label{admission-maturity-and-trust}

\textbf{Admission gates.} Before admission, Zep checks the verification of the evidence, the applicability, the conflicts, and the required evaluations. A candidate that fails a check is not admitted automatically and waits for a human to review it.

\textbf{Approval modes.} Each namespace uses \texttt{auto} or \texttt{manual} approval. In \texttt{auto} mode, Zep admits each candidate that passes the gates. In \texttt{manual} mode, a human must approve each candidate. A namespace can limit automatic approval to selected task families, and it can require a human to approve each Skill that uses selected tools. An approval must come from an authenticated human. An API credential cannot approve. Sharing a Skill with another namespace also requires a human. We chose \texttt{auto} as the default so that an admitted procedure can help on the next run. Manual approval would delay each candidate until a human reviewed it. The gates and the search weight (below) limit the risk of automatic admission. Verified failures lower the rank of a Skill, and a scheduled maintenance job can retire a provisional Skill that agents do not use. Search does not return a retired version.

\textbf{Evaluations.} Zep does not run application code, verifiers, or evaluations. Voyager \citep{voyager}, SkillRevise \citep{skillrevise}, and SkillLift \citep{skilllift} improve skills by execution of candidates. Execution would need a sandbox and the tool credentials of each application, which is impractical for a production service. A namespace can instead require evaluations for the candidates of a task family. The application reads the evaluations that each candidate needs, runs them in its own environment, and records each result on the candidate through the API. The record names the evaluator and its version, the environment, the verdict, and a reference to the evidence. Zep accepts a result only from a caller with access to the namespace. A failed evaluation rejects the result group, and the outcome history records the failure. A candidate without a passing result for each required evaluation stays a candidate. Scores from observability and evaluation platforms also enter as verification of the Traces themselves (Section 5.1).

\textbf{Group admission and versions.} A human reviewer sees the complete result group. Admission applies every change of the group in one transaction or applies none, so an approval cannot leave out a related conflict. Each admitted change creates an immutable Skill version. Earlier versions stay readable, and a restore of an earlier version creates a new version. SkillRevise \citep{skillrevise} also keeps versions and selects among them by measured utility.

\textbf{Maturity and trust.} Each admitted Skill version has a maturity tier (Table 2).

\par\noindent\begin{minipage}{\linewidth}
\needspace{8\baselineskip}
\textbf{Table 2.} Maturity tiers.\par\smallskip\noindent
\begin{tabular}{@{}
  >{\raggedright\arraybackslash}p{\dimexpr 0.2479\linewidth - 0.4958\tabcolsep\relax}
  >{\raggedright\arraybackslash}p{\dimexpr 0.7521\linewidth - 1.5042\tabcolsep\relax}@{}}
\toprule\noalign{}
\begin{minipage}[b]{\linewidth}\raggedright
Maturity
\end{minipage} & \begin{minipage}[b]{\linewidth}\raggedright
Condition
\end{minipage} \\
\midrule\noalign{}
\texttt{provisional} & One attempt family supports the Skill. \\
\texttt{corroborated} & Two or more attempt families support the Skill. \\
\texttt{established} & The trust value meets a configured threshold. \\
\bottomrule\noalign{}
\end{tabular}
\end{minipage}\par\medskip

A \emph{trial} is a full or partial use of a Skill whose outcome is a success or a failure with strong verification (Section 5.5 describes how an agent records a use). Other uses are not trials, for example a use in an abandoned attempt, a use with an unknown outcome, or a use whose outcome only the agent reports. The \emph{trust value} of a Skill is the lower bound of the 95\% Wilson score interval for its success rate over trials. The bound stays low until the Skill has enough trials. A Skill becomes \texttt{established} when its trust value meets the threshold.

\textbf{Search weight.} Search ranks a compiled Skill with a second value, the \emph{search weight}. The search weight is the posterior mean of the success rate. The prior uses the number of supporting attempt families \(n\) and a constant prior strength \(k\):

\[
w = \frac{k \cdot \frac{n}{n+1} + s}{k + t}
\]

Here, \(s\) is the number of successful trials and \(t\) is the total number of trials. The prior mean \(n/(n+1)\) increases with the number of supporting families, and \(k\) sets how many trials the prior is worth. Each trial changes \(w\), and verified failures can lower \(w\) below the prior mean. The trust value is conservative and controls maturity. The search weight changes with each trial and controls rank. Trials associate outcomes with Skill versions, but only a controlled evaluation can estimate the causal effect of a Skill.

\textbf{Single-source Skills.} Some procedures run rarely, so the first verified success can be the only evidence for a long time. A service that waits for several attempt families leaves the agent without the procedure until then. A service that admits a Skill from one attempt family risks a procedure that holds only for that instance, for example a step that depends on a value of that task. Zep admits a \emph{single-source Skill}, which is a Skill that one attempt family supports, and it limits the risk in three ways. First, the compiler applies the stricter checks of Section 5.3: the procedure must not keep identifiers from the source, and its applicability must stay within the tools and environments of the evidence. Second, the Skill starts as \texttt{provisional} and has a lower search weight than a Skill with more support. Third, when later attempt families support a compatible procedure, the Curator patches or merges the Skill, and its maturity increases.

\needspace{10\baselineskip}
\subsection{Retrieval and use}\label{retrieval-and-use}

An agent searches with an objective. The request can also include optional filters, such as a task family, tools, and an environment. An application that declares the task family of its Traces (Section 5.1) can pass the same family to search. Search first keeps only the active Skills of the namespace that the caller may read and whose applicability fields match the request. It excludes candidates, retired versions, and Skills with an unresolved conflict. Relevance cannot override these filters.

Search then matches the objective against a full-text index of a short \emph{descriptor} of each Skill: its name, its description, and its applicability. Search ranks each result by its lexical score times its search weight (Section 5.4). The search weight changes the order of eligible Skills, and it cannot make an ineligible Skill visible.

Retrieval has two steps, as in the progressive disclosure of the Agent Skills format \citep{agentskills, claudeskills}. First, search returns the descriptors of the matching Skills. The agent reviews the descriptors and selects the Skills that apply to its task. Second, the agent reads the body of each Skill that it selected. The agent thus loads a full procedure only for a Skill that it will use.

Retrieval alone is not evidence that the agent used a Skill. The agent harness records a \emph{Skill use}, which links a Skill version to the Trace of the task. The use outcome inherits the verification of the Trace only when the two outcomes match and the verifier is still registered. Uses with a strongly verified outcome are the trials of Section 5.4.

We did not use vector retrieval. A descriptor is short, curated text with explicit names and use conditions, and we did not expect embedding similarity to select Skills better than lexical search on this text. Topic similarity can also return a Skill whose preconditions do not match the task. The experiments do not compare the two methods. A survey of dynamic skill libraries \citep{dynamicskills} reports that selection from a flat store, which has no hierarchy or filter, can degrade when a library reaches tens to hundreds of skills, because each new skill adds distractors. The structured filters of Zep Agent Skills reduce the set of descriptors that the agent reviews.

\needspace{10\baselineskip}
\subsection{Governance and safety}\label{governance-and-safety}

A poisoned Skill is more dangerous than a poisoned memory record, because the agent follows a Skill as instructions (Section 3.2). Zep Agent Skills can admit a Skill without human review (Section 5.4), so each control is an automatic check. Table 3 maps each threat of Section 3.2 to its controls. Most controls are part of the stages above. This section adds the controls for access, sharing, and deletion.

\par\noindent\begin{minipage}{\linewidth}
\needspace{8\baselineskip}
\textbf{Table 3.} Threats and controls.\par\smallskip\noindent
\begin{tabular}{@{}
  >{\raggedright\arraybackslash}p{\dimexpr 0.3051\linewidth - 0.6102\tabcolsep\relax}
  >{\raggedright\arraybackslash}p{\dimexpr 0.6949\linewidth - 1.3898\tabcolsep\relax}@{}}
\toprule\noalign{}
\begin{minipage}[b]{\linewidth}\raggedright
Threat
\end{minipage} & \begin{minipage}[b]{\linewidth}\raggedright
Controls
\end{minipage} \\
\midrule\noalign{}
Instance-data leakage & The Reflector replaces instance values with placeholders, and the scanners reject candidates that contain instance data (Section 5.3). \\
Instruction injection & The LLM roles read Trace content as data, and the scanners reject instructions that come from tool output (Section 5.3). \\
Unverified or poisoned outcomes & Zep sets the verification strength from the identity of the verifier, and only strongly verified successes support a new Skill (Sections 5.1 and 5.3). \\
Cross-boundary flow & Each Skill comes from the evidence of one namespace, and restricted events are not compiled (Section 5.3). The access control and sharing rules below also apply. \\
Unsafe procedures & Zep does not run Skills. A namespace can require human approval for each Skill that uses selected tools (Section 5.4). \\
\bottomrule\noalign{}
\end{tabular}
\end{minipage}\par\medskip

\textbf{Access control.} Each namespace belongs to one security domain, for example one organization or one application. Access control applies the same rules to capture, compilation, admission, retrieval, and MCP delivery. An agent can search and read only the Skills of its own namespace.

\textbf{Sharing.} Namespaces do not share Skills by default. A human can publish a Skill to another namespace in the same security domain. The copy is then a candidate of the destination namespace, and it becomes searchable only after the admission policy of that namespace accepts it.

\textbf{Deletion.} When a human deletes a source Trace or event, each Skill that cites it becomes unavailable at once, so a procedure that depends on deleted evidence does not stay searchable. Zep then schedules a repair of each affected Skill, and an asynchronous job removes the deleted content.

\textbf{Audit.} Zep keeps an audit record of each verification, admission, and publication decision.

\needspace{10\baselineskip}
\section{Evaluation method}\label{evaluation-method}

\needspace{10\baselineskip}
\subsection{Research questions}\label{research-questions}

The evaluation answers six questions:

\begin{itemize}
\tightlist
\item
  \textbf{RQ1.} Do Skills that Zep compiles from teacher Traces increase the pass rate of an executor agent on repeated tasks?
\item
  \textbf{RQ2.} Does a higher reasoning effort give the same gain as Skills?
\item
  \textbf{RQ3.} Do Skills change the reliability over repeated trials?
\item
  \textbf{RQ4.} Do Skills change the tokens and the latency for each passed task?
\item
  \textbf{RQ5.} Which tasks get the gain: tasks that the teacher solved, or tasks that the teacher did not solve?
\item
  \textbf{RQ6.} How does the gain from Skills compare with the gain from continuous fine-tuning of the model weights?
\end{itemize}

The main experiments do not measure transfer to tasks that the compiler did not see, because of the design of the benchmarks. In \ensuremath{\tau}\ensuremath{^3} and EnterpriseOps-Gym, most tasks share little similarity with the other tasks of the benchmark, and in AppWorld only the variants of one scenario are similar. A split of these benchmarks into train and test tasks would leave few test tasks with a procedure that a train task teaches. Section 7.5 gives the limited transfer evidence that the experiments contain.

\needspace{10\baselineskip}
\subsection{Protocol}\label{protocol}

Each main benchmark uses the same five stages:

\begin{enumerate}
\def\labelenumi{\arabic{enumi}.}
\tightlist
\item
  \textbf{Teach.} A teacher agent runs each task. The harness records each run as a Trace with ordered events, an outcome, and external verification from the benchmark checker.
\item
  \textbf{Compile.} The harness sends the Traces to a Zep namespace. The production compiler (Section 5.3) compiles Skills from the Traces. Admission is automatic.
\item
  \textbf{Freeze.} The harness exports the active Skills to a snapshot file and records the SHA-256 digest of the snapshot. The executors read only the snapshot, so continued compilation cannot change an arm.
\item
  \textbf{Execute.} Each executor runs each task under two conditions, the two \emph{arms} of the experiment. The \emph{no-Skills arm} is the baseline and has the benchmark tools only. The \emph{Skills arm} has the same tools, a Skill index in the system prompt, and a \texttt{read\_skill} tool that returns the Markdown body of one Skill from the snapshot. The Skills arm thus differs from the baseline in three ways: the Skill content, a longer system prompt, and one more tool. The experiments do not separate these effects (Section 7.4).
\item
  \textbf{Compare.} The analysis pairs the two arms on the same task and trial.
\end{enumerate}

The teacher-executor configuration also models a deployment pattern for production. An application can send a task that it has not seen before to a stronger model. Zep Agent Skills compiles the verified successes of that model into Skills, and a cheaper and faster model then uses these Skills when the task occurs again. The experiments measure the second step of this pattern: the gain of the cheaper executors from the Skills of the teacher.

The Skills arm uses progressive disclosure (Section 5.5), but the harness does not call search. It puts the Skill index in the system prompt. The index holds the name and the description of each Skill. The agent selects a Skill from the index and reads its body with \texttt{read\_skill}. The harness does not insert a Skill body into the prompt.

A cell is one executor model at one reasoning effort on one benchmark. The two arms of a cell run at the same time, because model APIs and serving change between runs. A paired interval within one run does not include this drift between runs. Different cells ran at different times and at different concurrency, so a latency comparison is valid only between the two arms of one cell.

The scope of the Skill index is different for each benchmark (Table 4). When a request names a task family, search returns only the Skills of that family (the family index). The EnterpriseOps-Gym harness used the family index. The \ensuremath{\tau}\ensuremath{^3} and AppWorld harnesses gave the agent the Skills of all task families, so the agent had to select from a larger catalog than a family search gives. This catalog is the flat store of Section 5.5.

\par\noindent\begin{minipage}{\linewidth}
\needspace{8\baselineskip}
\textbf{Table 4. Skill delivery in each benchmark.}\par\smallskip\noindent
\begin{tabular}{@{}
  >{\raggedright\arraybackslash}p{\dimexpr 0.2629\linewidth - 1.5774\tabcolsep\relax}
  >{\raggedleft\arraybackslash}p{\dimexpr 0.1429\linewidth - 0.8574\tabcolsep\relax}
  >{\raggedleft\arraybackslash}p{\dimexpr 0.1026\linewidth - 0.6156\tabcolsep\relax}
  >{\raggedright\arraybackslash}p{\dimexpr 0.4915\linewidth - 2.9490\tabcolsep\relax}@{}}
\toprule\noalign{}
\begin{minipage}[b]{\linewidth}\raggedright
Benchmark
\end{minipage} & \begin{minipage}[b]{\linewidth}\raggedleft
Skills in the snapshot
\end{minipage} & \begin{minipage}[b]{\linewidth}\raggedleft
Task families
\end{minipage} & \begin{minipage}[b]{\linewidth}\raggedright
Index scope in the Skills arm
\end{minipage} \\
\midrule\noalign{}
\ensuremath{\tau}\ensuremath{^3}-bench banking knowledge & 31 & --- & All Skills \\
AppWorld & 62 & 48 & All Skills \\
EnterpriseOps-Gym & 275 & 8 & Skills of the task family (mean 44.3 in the index) \\
ALFWorld & 22 & 6 & The top 3 search results, inserted as bodies \\
\bottomrule\noalign{}
\end{tabular}
\end{minipage}\par\medskip

\needspace{10\baselineskip}
\subsection{Benchmarks}\label{benchmarks}

We chose the three main benchmarks because their tasks are long-running business tasks. Together, they reflect the breadth of the tasks that Zep customers do: customer support for a bank, where the agent searches a knowledge base and changes accounts with tools (\ensuremath{\tau}\ensuremath{^3}), code that calls the APIs of applications (AppWorld), and enterprise operations under written policies (EnterpriseOps-Gym). \ensuremath{\tau}\ensuremath{^3} and EnterpriseOps-Gym are also challenging for today's frontier models and not saturated: the teacher passes 57.2\% and 52.8\% of their tasks, respectively (Section 7.2). We chose ALFWorld because it shows how Skills compare with continuous fine-tuning: the RSD paper \citep{rsd} reports its results on this benchmark.

\par\noindent\begin{minipage}{\linewidth}
\needspace{8\baselineskip}
\textbf{Table 5. Benchmarks.}\par\smallskip\noindent
\begin{tabular}{@{}
  >{\raggedright\arraybackslash}p{\dimexpr 0.2629\linewidth - 1.5774\tabcolsep\relax}
  >{\raggedleft\arraybackslash}p{\dimexpr 0.3081\linewidth - 1.8486\tabcolsep\relax}
  >{\raggedleft\arraybackslash}p{\dimexpr 0.1208\linewidth - 0.7248\tabcolsep\relax}
  >{\raggedright\arraybackslash}p{\dimexpr 0.3081\linewidth - 1.8486\tabcolsep\relax}@{}}
\toprule\noalign{}
\begin{minipage}[b]{\linewidth}\raggedright
Benchmark
\end{minipage} & \begin{minipage}[b]{\linewidth}\raggedleft
Tasks
\end{minipage} & \begin{minipage}[b]{\linewidth}\raggedleft
Trials per task and arm
\end{minipage} & \begin{minipage}[b]{\linewidth}\raggedright
Task family
\end{minipage} \\
\midrule\noalign{}
\ensuremath{\tau}\ensuremath{^3}-bench banking knowledge & 97 & 4 & The type of the first write action in the expected solution \\
AppWorld & 147 (49 scenarios \ensuremath{\times} 3 variants) & 3 & The scenario \\
EnterpriseOps-Gym & 625 & 3 & The domain (8) \\
ALFWorld & 134 \texttt{valid\_unseen} games & 3 & The task type (6) \\
\bottomrule\noalign{}
\end{tabular}
\end{minipage}\par\medskip

\textbf{\ensuremath{\tau}\ensuremath{^3}-bench banking knowledge.} An agent serves a simulated bank customer with banking tools and a knowledge base. Some tools are \emph{discoverable}: the agent does not see them in its tool list, and it must find their names in the knowledge base. Each such name has a numeric suffix that the agent cannot guess. The user simulator is GPT-5.2 at low reasoning effort. Each episode has a limit of 200 steps.

\textbf{AppWorld.} An agent writes Python code against the APIs of nine simulated applications. Each scenario has three variants with different values. The harness runs the Train and Dev scenarios. An episode has a limit of 40 steps. Earlier research often used AppWorld, but the benchmark is now saturated: the teacher passes 97.3\% of its tasks (Section 7.2). We include it for comparison with earlier research.

\textbf{EnterpriseOps-Gym.} An agent does enterprise operations through MCP tools in eight domains: calendar, customer service management (CSM), drive, email, HR, hybrid, IT service management (ITSM), and Teams. A written policy controls each domain. SQL checks on the final database state score each task.

\textbf{ALFWorld.} An agent completes household tasks in a text environment. The experiment uses the setup of the RSD paper \citep{rsd}, so that the paper can compare Skills with continuous fine-tuning. Section 6.6 describes the differences of this experiment.

\textbf{SkillsBench.} We mentioned SkillsBench \citep{skillsbench} earlier in Section 2. We do not use it, because its Skills arm is not comparable with ours. SkillsBench supplies a curated Skill for each task. Zep Agent Skills compiles Skills only from the runs that the teacher passed.

Appendix A lists each place where the experiments deviate from the original method, dataset, or harness of a benchmark. The benchmark corrections change both arms. In EnterpriseOps-Gym, the corrected score is 3.6 to 5.5 points higher than the original score in each arm. The Skills gain on the original score is +7.9 to +11.0 points, and on the corrected score +8.4 to +11.4 points.

\needspace{10\baselineskip}
\subsection{Models, metrics, and statistics}\label{models-metrics-and-statistics}

\par\noindent\begin{minipage}{\linewidth}
\needspace{8\baselineskip}
\textbf{Table 6. Models.}\par\smallskip\noindent
\begin{tabular}{@{}
  >{\raggedright\arraybackslash}p{\dimexpr 0.1887\linewidth - 1.1322\tabcolsep\relax}
  >{\raggedright\arraybackslash}p{\dimexpr 0.2188\linewidth - 1.3128\tabcolsep\relax}
  >{\raggedright\arraybackslash}p{\dimexpr 0.3381\linewidth - 2.0286\tabcolsep\relax}
  >{\raggedright\arraybackslash}p{\dimexpr 0.2543\linewidth - 1.5258\tabcolsep\relax}@{}}
\toprule\noalign{}
\begin{minipage}[b]{\linewidth}\raggedright
Role
\end{minipage} & \begin{minipage}[b]{\linewidth}\raggedright
Model
\end{minipage} & \begin{minipage}[b]{\linewidth}\raggedright
Benchmarks
\end{minipage} & \begin{minipage}[b]{\linewidth}\raggedright
Reasoning effort
\end{minipage} \\
\midrule\noalign{}
Teacher & Claude Opus 5.5 & \ensuremath{\tau}\ensuremath{^3}, AppWorld, EnterpriseOps-Gym & \texttt{max} \\
Executor & GPT-6 Luna & \ensuremath{\tau}\ensuremath{^3}, AppWorld, EnterpriseOps-Gym & \texttt{low}, \texttt{medium}, \texttt{max} \\
Executor & Claude Haiku 5.5 & \ensuremath{\tau}\ensuremath{^3}, AppWorld, EnterpriseOps-Gym & \texttt{low}, \texttt{medium}, \texttt{max} \\
Executor & DeepSeek V4.1 Flash & \ensuremath{\tau}\ensuremath{^3}, AppWorld, EnterpriseOps-Gym & \texttt{low}, \texttt{high}, \texttt{max} (\ensuremath{\tau}\ensuremath{^3}: \texttt{low}, \texttt{max}) \\
Teacher, Executor & Qwen3-8B & ALFWorld & Non-thinking mode \\
\bottomrule\noalign{}
\end{tabular}
\end{minipage}\par\medskip

\textbf{Metrics.}

\begin{itemize}
\tightlist
\item
  \textbf{Pass rate.} The share of episodes that pass the benchmark checker. An empty response from the inference provider, a context overflow, and a step-limit stop count as failures.
\item
  \textbf{pass\ensuremath{^4}.} The share of \ensuremath{\tau}\ensuremath{^3} tasks that pass in all four trials.
\item
  \textbf{Scenario pass rate.} The share of AppWorld scenarios with all three variants passed in one trial.
\item
  \textbf{Tokens for each passed episode.} All executor tokens of a cell divided by the passed episodes of the cell. The value includes the tokens of failed episodes. The value excludes the user simulator and the compiler.
\item
  \textbf{Latency.} The mean end-to-end time of an episode. \ensuremath{\tau}\ensuremath{^3} uses the passed episodes, and AppWorld uses all episodes. The harnesses ran in the data centers that Cognition uses for Devin, so each latency includes the network path from these data centers to the model API.
\item
  \textbf{Read rate.} The share of Skills-arm episodes with at least one \texttt{read\_skill} call.
\end{itemize}

\textbf{Statistics.} Each comparison pairs the two arms on the same task and trial. The reports use these methods:

\par\noindent\begin{minipage}{\linewidth}
\par\smallskip\noindent
\begin{tabular}{@{}
  >{\raggedright\arraybackslash}p{\dimexpr 0.1932\linewidth - 0.7728\tabcolsep\relax}
  >{\raggedright\arraybackslash}p{\dimexpr 0.4034\linewidth - 1.6136\tabcolsep\relax}
  >{\raggedright\arraybackslash}p{\dimexpr 0.4034\linewidth - 1.6136\tabcolsep\relax}@{}}
\toprule\noalign{}
\begin{minipage}[b]{\linewidth}\raggedright
Benchmark
\end{minipage} & \begin{minipage}[b]{\linewidth}\raggedright
Interval
\end{minipage} & \begin{minipage}[b]{\linewidth}\raggedright
Test and multiple-comparison correction
\end{minipage} \\
\midrule\noalign{}
\ensuremath{\tau}\ensuremath{^3} & Paired bootstrap over tasks, 100,000 draws, all trials of a task together & Paired randomization test; Holm correction over 8 cells \\
AppWorld & Bootstrap over scenarios, 4,000 draws & Bonferroni 99.44\% intervals for 9 comparisons \\
EnterpriseOps-Gym & Bootstrap over tasks, 10,000 draws & 95\% intervals; no test \\
ALFWorld & Paired bootstrap over games & 95\% intervals; no test \\
\bottomrule\noalign{}
\end{tabular}
\end{minipage}\par\medskip

The teacher strata in Section 7.5 are exploratory. They have no separate multiplicity correction.

\needspace{10\baselineskip}
\subsection{Isolation and leakage controls}\label{isolation-and-leakage-controls}

A Skill must carry a procedure and must not carry the answer of a scored task. The harnesses enforce these controls:

\begin{itemize}
\tightlist
\item
  \textbf{Tool isolation.} The executor gets only the benchmark tools, \texttt{read\_skill}, and a finish tool. The executors have no web access.
\item
  \textbf{Teacher isolation.} The EnterpriseOps-Gym teacher runs in Claude Code with the shell, file, web fetch, and web search tools disabled.
\item
  \textbf{Hidden checks.} No agent sees the checker SQL, the reference solutions, or the database IDs.
\item
  \textbf{Leakage scans.} A scan of the EnterpriseOps-Gym Skills looked for checker SQL, database IDs, and values from the hidden checks. It found none, other than API field names, common words, and one HR service name that the HR tools return to the agent. In AppWorld, no executor code read the answers or the test data.
\end{itemize}

The \ensuremath{\tau}\ensuremath{^3} Skills contain the names of discoverable tools. The experiment does not separate the effect of these names from the effect of the procedures.

\needspace{10\baselineskip}
\subsection{ALFWorld setup}\label{alfworld-setup}

The ALFWorld experiment differs from the main experiments in four ways:

\begin{enumerate}
\def\labelenumi{\arabic{enumi}.}
\tightlist
\item
  \textbf{Skill delivery.} The harness inserted the top three search results as Skill bodies. The harness did not use the index and \texttt{read\_skill}.
\item
  \textbf{Self-teaching.} Qwen3-8B is the teacher and the executor. The teacher ran 34 train games of each of the six task types (204 Traces), and Zep compiled 22 Skills. RSD also learns from the runs of the same model.
\item
  \textbf{Weak model.} Qwen3-8B is a small model. We used it because the RSD paper reports its results on Qwen3-8B, so we can compare the two methods on the same model.
\item
  \textbf{No controlled comparison with RSD.} We rebuilt the RSD harness from the paper text (Appendix A). Thus the comparison uses the gain of each method over its own base.
\end{enumerate}

Each ALFWorld game is a household task in one of six task types, for example ``put an object in a place'' or ``heat an object and put it in a place''. The prompt of each game includes one demonstration: the full steps of a different game that is solved. The choice of the demonstration changes the base score by about 32 points. The experiment reports two setups:

\begin{itemize}
\tightlist
\item
  \textbf{\texttt{fixed}:} each game gets the same demonstration, a solved ``put an object in a place'' game. Thus, a heat game gets no example of how to heat an object. This setup is the weak base (42.8\%).
\item
  \textbf{\texttt{typed}:} each game gets a demonstration of its own task type. For example, a heat game gets a solved heat game. This setup is the strong base (74.6\%).
\end{itemize}

The RSD paper reports 62.9\% for Qwen3-8B before RSD training. This score is between the scores of the two setups.

\needspace{10\baselineskip}
\section{Results}\label{results}

\needspace{10\baselineskip}
\subsection{Summary of findings}\label{summary-of-findings}

Compiled Skills increase the pass rate of each executor model at each reasoning effort, by 4.5 to 18.8 points. Most of the gain comes from the tasks that the teacher solved, so the success of the teacher limits the gain. Skills also make the executors more reliable across trials, and in most model and benchmark pairs, a higher reasoning effort does not replace them.

\par\noindent\begin{minipage}{\linewidth}
\needspace{8\baselineskip}
\textbf{Table 7. Findings.}\par\smallskip\noindent
\begin{tabular}{@{}
  >{\raggedright\arraybackslash}p{\dimexpr 0.0349\linewidth - 0.2094\tabcolsep\relax}
  >{\raggedright\arraybackslash}p{\dimexpr 0.3217\linewidth - 1.9302\tabcolsep\relax}
  >{\raggedright\arraybackslash}p{\dimexpr 0.3217\linewidth - 1.9302\tabcolsep\relax}
  >{\raggedright\arraybackslash}p{\dimexpr 0.3217\linewidth - 1.9302\tabcolsep\relax}@{}}
\toprule\noalign{}
\begin{minipage}[b]{\linewidth}\raggedright
\#
\end{minipage} & \begin{minipage}[b]{\linewidth}\raggedright
Finding
\end{minipage} & \begin{minipage}[b]{\linewidth}\raggedright
Evidence
\end{minipage} & \begin{minipage}[b]{\linewidth}\raggedright
Limit
\end{minipage} \\
\midrule\noalign{}
F1 & Skills increase the pass rate on repeated tasks for each executor model and each effort. & 26 of 26 cells in three benchmarks have a positive paired difference (Tables 8 to 10). & Replay design: the executors run the tasks that the teacher ran. \\
F2 & The gain is largest for the tasks that the teacher solved. On tasks that the teacher did not solve, the gain is small or 0. & Teacher strata in \ensuremath{\tau}\ensuremath{^3} and EnterpriseOps-Gym, and source tasks against sibling tasks in AppWorld (Section 7.5). & The strata are exploratory. \\
F3 & A higher reasoning effort does not replace Skills in most model and benchmark pairs. & In EnterpriseOps-Gym, Skills at \texttt{low} effort pass more tasks than no Skills at \texttt{max} effort for all three models (Section 7.3). & The result holds in 2 of 3 models on AppWorld and 1 of 3 on \ensuremath{\tau}\ensuremath{^3}. \\
F4 & Skills increase reliability. & \ensuremath{\tau}\ensuremath{^3} pass\ensuremath{^4} increases in 7 of 8 cells. AppWorld scenario pass rate increases in 9 of 9 cells (Section 7.3). & Three or four trials for each task. \\
F5 & The Skill index adds a fixed cost to each model call. Long tasks amortize this cost: their tokens for each passed episode decrease. On short tasks, the cost is larger than the saving. The pass rate increases on both. & \ensuremath{\tau}\ensuremath{^3} (long episodes): \ensuremath{-}7.0\% to \ensuremath{-}44.6\% in 8 of 8 cells. AppWorld (short episodes): an increase in 7 of 9 cells, and a decrease in the 2 cells with the longest episodes (Section 7.4). & An interpretation: no arm isolates the cost of the index. The AppWorld index held all 62 Skills (Table 4). A production agent uses search, not an index (Section 7.4). \\
F6 & Skills from the runs of a weak model give a large gain without a change to the weights. The gain is similar to or larger than the gain that RSD reports. & ALFWorld \texttt{valid\_unseen}: +32.6 points from the weak base and +8.2 from the strong base, against +13.7 points for RSD (Section 7.6). & The bases and the harnesses differ, so the comparison uses gains. \\
\bottomrule\noalign{}
\end{tabular}
\end{minipage}\par\medskip

\needspace{10\baselineskip}
\subsection{Main results}\label{main-results}

\par\noindent\begin{minipage}{\linewidth}
\needspace{8\baselineskip}
\textbf{Table 8. \ensuremath{\tau}\ensuremath{^3}-bench banking knowledge.} 97 tasks \ensuremath{\times} 4 trials = 388 episodes for each arm. The teacher row is the pass rate of the teacher (Claude Opus 5.5, maximum effort, no Skills) over the same 388 episodes.\par\smallskip\noindent
\begin{tabular}{@{}
  >{\raggedright\arraybackslash}p{\dimexpr 0.2745\linewidth - 2.7450\tabcolsep\relax}
  >{\raggedright\arraybackslash}p{\dimexpr 0.0797\linewidth - 0.7970\tabcolsep\relax}
  >{\raggedleft\arraybackslash}p{\dimexpr 0.1088\linewidth - 1.0880\tabcolsep\relax}
  >{\raggedleft\arraybackslash}p{\dimexpr 0.0797\linewidth - 0.7970\tabcolsep\relax}
  >{\raggedleft\arraybackslash}p{\dimexpr 0.2877\linewidth - 2.8770\tabcolsep\relax}
  >{\raggedleft\arraybackslash}p{\dimexpr 0.1696\linewidth - 1.6960\tabcolsep\relax}@{}}
\toprule\noalign{}
\begin{minipage}[b]{\linewidth}\raggedright
Executor
\end{minipage} & \begin{minipage}[b]{\linewidth}\raggedright
Effort
\end{minipage} & \begin{minipage}[b]{\linewidth}\raggedleft
No Skills
\end{minipage} & \begin{minipage}[b]{\linewidth}\raggedleft
Skills
\end{minipage} & \begin{minipage}[b]{\linewidth}\raggedleft
Difference, points {[}95\% CI{]}
\end{minipage} & \begin{minipage}[b]{\linewidth}\raggedleft
Holm-adjusted p
\end{minipage} \\
\midrule\noalign{}
GPT-6 Luna & low & 13.1\% & 21.6\% & +8.5 {[}+3.4, +13.9{]} & 0.022 \\
GPT-6 Luna & medium & 30.2\% & 36.1\% & +5.9 {[}+0.5, +11.6{]} & 0.117 \\
GPT-6 Luna & max & 38.9\% & 44.1\% & +5.2 {[}+0.5, +9.8{]} & 0.117 \\
Claude Haiku 5.5 & low & 19.1\% & 28.9\% & +9.8 {[}+3.1, +16.5{]} & 0.036 \\
Claude Haiku 5.5 & medium & 21.1\% & 36.6\% & +15.5 {[}+9.5, +21.7{]} & 0.0002 \\
Claude Haiku 5.5 & max & 43.3\% & 51.8\% & +8.5 {[}+2.3, +15.0{]} & 0.051 \\
DeepSeek V4.1 Flash & low & 42.0\% & 50.8\% & +8.8 {[}+3.1, +14.7{]} & 0.036 \\
DeepSeek V4.1 Flash & max & 46.4\% & 52.8\% & +6.4 {[}+0.3, +12.9{]} & 0.117 \\
Teacher: Claude Opus 5.5 & max & 57.2\% & --- & --- & --- \\
\bottomrule\noalign{}
\end{tabular}
\end{minipage}\par\medskip

All eight unadjusted intervals are above 0. Four of eight cells pass the Holm threshold at a family-wise \ensuremath{\alpha} of 0.05. The pooled gain over the eight cells is +8.6 points (95\% CI +5.5 to +11.9). An empty response from the inference provider counts as a failure. A sensitivity view that removes each pair with an empty response gives the same direction in each cell.

\par\noindent\begin{minipage}{\linewidth}
\needspace{8\baselineskip}
\textbf{Table 9. AppWorld, Train and Dev.} 147 tasks \ensuremath{\times} 3 trials = 441 episodes for each arm. The teacher row is the pass rate of the teacher run, one trial for each task (143 of 147 tasks).\par\smallskip\noindent
\begin{tabular}{@{}
  >{\raggedright\arraybackslash}p{\dimexpr 0.2831\linewidth - 3.3972\tabcolsep\relax}
  >{\raggedright\arraybackslash}p{\dimexpr 0.0822\linewidth - 0.9864\tabcolsep\relax}
  >{\raggedleft\arraybackslash}p{\dimexpr 0.0849\linewidth - 1.0188\tabcolsep\relax}
  >{\raggedleft\arraybackslash}p{\dimexpr 0.0822\linewidth - 0.9864\tabcolsep\relax}
  >{\raggedleft\arraybackslash}p{\dimexpr 0.2262\linewidth - 2.7144\tabcolsep\relax}
  >{\raggedleft\arraybackslash}p{\dimexpr 0.1426\linewidth - 1.7112\tabcolsep\relax}
  >{\raggedleft\arraybackslash}p{\dimexpr 0.0989\linewidth - 1.1868\tabcolsep\relax}@{}}
\toprule\noalign{}
\begin{minipage}[b]{\linewidth}\raggedright
Executor
\end{minipage} & \begin{minipage}[b]{\linewidth}\raggedright
Effort
\end{minipage} & \begin{minipage}[b]{\linewidth}\raggedleft
No Skills
\end{minipage} & \begin{minipage}[b]{\linewidth}\raggedleft
Skills
\end{minipage} & \begin{minipage}[b]{\linewidth}\raggedleft
Difference, points {[}95\% CI{]}
\end{minipage} & \begin{minipage}[b]{\linewidth}\raggedleft
99.44\% CI
\end{minipage} & \begin{minipage}[b]{\linewidth}\raggedleft
Wins / losses
\end{minipage} \\
\midrule\noalign{}
GPT-6 Luna & low & 68.5\% & 87.3\% & +18.8 {[}+10.2, +28.3{]} & {[}+6.1, +32.0{]} & 108 / 25 \\
GPT-6 Luna & medium & 86.8\% & 94.6\% & +7.7 {[}+2.7, +13.4{]} & {[}+1.4, +15.6{]} & 42 / 8 \\
GPT-6 Luna & max & 91.4\% & 96.6\% & +5.2 {[}+1.4, +10.2{]} & {[}+0.2, +12.7{]} & 26 / 3 \\
Claude Haiku 5.5 & low & 88.0\% & 94.8\% & +6.8 {[}+0.9, +13.6{]} & {[}\ensuremath{-}1.4, +16.3{]} & 41 / 11 \\
Claude Haiku 5.5 & medium & 91.6\% & 97.1\% & +5.4 {[}+1.4, +10.9{]} & {[}+0.2, +13.2{]} & 27 / 3 \\
Claude Haiku 5.5 & max & 92.7\% & 98.0\% & +5.2 {[}+1.1, +10.7{]} & {[}+0.5, +13.2{]} & 23 / 0 \\
DeepSeek V4.1 Flash & low & 92.7\% & 97.3\% & +4.5 {[}+1.8, +7.7{]} & {[}+0.9, +9.1{]} & 21 / 1 \\
DeepSeek V4.1 Flash & high & 91.6\% & 97.3\% & +5.7 {[}+1.6, +10.7{]} & {[}+0.5, +13.4{]} & 28 / 3 \\
DeepSeek V4.1 Flash & max & 91.8\% & 96.8\% & +5.0 {[}+2.5, +7.9{]} & {[}+1.6, +9.3{]} & 24 / 2 \\
Teacher: Claude Opus 5.5 & max & 97.3\% & --- & --- & --- & --- \\
\bottomrule\noalign{}
\end{tabular}
\end{minipage}\par\medskip

Eight of nine Bonferroni intervals are above 0. The Claude Haiku 5.5 \texttt{low} interval includes 0. The no-Skills pass rates are high (68.5\% to 92.7\%), and the teacher passes 97.3\%, so the room for a gain is small in most cells. These rates show the saturation of AppWorld that Section 6.3 mentions.

\par\noindent\begin{minipage}{\linewidth}
\needspace{8\baselineskip}
\textbf{Table 10. EnterpriseOps-Gym.} 625 tasks \ensuremath{\times} 3 trials = 1,875 episodes for each arm. Corrected score (Appendix A). The teacher row is the pass rate of the teacher run, one trial for each task (330 of 625 tasks).\par\smallskip\noindent
\begin{tabular}{@{}
  >{\raggedright\arraybackslash}p{\dimexpr 0.2745\linewidth - 2.7450\tabcolsep\relax}
  >{\raggedright\arraybackslash}p{\dimexpr 0.0797\linewidth - 0.7970\tabcolsep\relax}
  >{\raggedleft\arraybackslash}p{\dimexpr 0.1116\linewidth - 1.1160\tabcolsep\relax}
  >{\raggedleft\arraybackslash}p{\dimexpr 0.0797\linewidth - 0.7970\tabcolsep\relax}
  >{\raggedleft\arraybackslash}p{\dimexpr 0.2964\linewidth - 2.9640\tabcolsep\relax}
  >{\raggedleft\arraybackslash}p{\dimexpr 0.1580\linewidth - 1.5800\tabcolsep\relax}@{}}
\toprule\noalign{}
\begin{minipage}[b]{\linewidth}\raggedright
Executor
\end{minipage} & \begin{minipage}[b]{\linewidth}\raggedright
Effort
\end{minipage} & \begin{minipage}[b]{\linewidth}\raggedleft
No Skills
\end{minipage} & \begin{minipage}[b]{\linewidth}\raggedleft
Skills
\end{minipage} & \begin{minipage}[b]{\linewidth}\raggedleft
Difference, points {[}95\% CI{]}
\end{minipage} & \begin{minipage}[b]{\linewidth}\raggedleft
Tasks up / down
\end{minipage} \\
\midrule\noalign{}
GPT-6 Luna & low & 27.7\% & 37.9\% & +10.2 {[}+7.8, +12.6{]} & 154 / 43 \\
GPT-6 Luna & medium & 31.8\% & 41.6\% & +9.8 {[}+7.5, +12.1{]} & 144 / 42 \\
GPT-6 Luna & max & 33.3\% & 42.5\% & +9.1 {[}+6.9, +11.4{]} & 134 / 37 \\
Claude Haiku 5.5 & low & 33.9\% & 45.3\% & +11.4 {[}+8.8, +14.1{]} & 149 / 47 \\
Claude Haiku 5.5 & medium & 35.7\% & 45.9\% & +10.1 {[}+7.6, +12.6{]} & 144 / 48 \\
Claude Haiku 5.5 & max & 35.8\% & 47.2\% & +11.4 {[}+8.9, +13.9{]} & 137 / 33 \\
DeepSeek V4.1 Flash & low & 48.3\% & 56.8\% & +8.5 {[}+6.0, +10.9{]} & 156 / 70 \\
DeepSeek V4.1 Flash & high & 48.4\% & 57.1\% & +8.6 {[}+6.1, +11.2{]} & 153 / 61 \\
DeepSeek V4.1 Flash & max & 49.9\% & 58.3\% & +8.4 {[}+6.1, +10.7{]} & 146 / 56 \\
Teacher: Claude Opus 5.5 & max & 52.8\% & --- & --- & --- \\
\bottomrule\noalign{}
\end{tabular}
\end{minipage}\par\medskip

All nine intervals are above 0. The gain is +8.4 to +11.4 points, and it changes little with the effort. On the original score, the gain is +7.9 to +11.0 points. For GPT-6 Luna and Claude Haiku 5.5, the point estimate is positive in each of the eight domains. For DeepSeek V4.1 Flash, the calendar, email, and Teams intervals include 0 at each effort.

The teacher limits the gain. The teacher is Claude Opus 5.5 at maximum effort, and it passes more tasks than each executor without Skills on each benchmark. A Skill of a task comes from a verified success of the teacher on that task, so a task that the teacher failed has no Skill of its own (Section 7.5). The Skills arm moves each executor toward the teacher: from 13.1\%--46.4\% to 21.6\%--52.8\% against 57.2\% on \ensuremath{\tau}\ensuremath{^3}, and from 68.5\%--92.7\% to 87.3\%--98.0\% against 97.3\% on AppWorld. The limit is not strict. On EnterpriseOps-Gym, DeepSeek V4.1 Flash with Skills passes 56.8\% to 58.3\% against 52.8\% for the teacher, because the teacher ran each task one time and the executors also gain on tasks that the teacher failed.

Figure 4 shows the paired differences of all 26 cells.

\begin{figure}
\centering
\includegraphics[width=1\linewidth,height=\textheight,keepaspectratio]{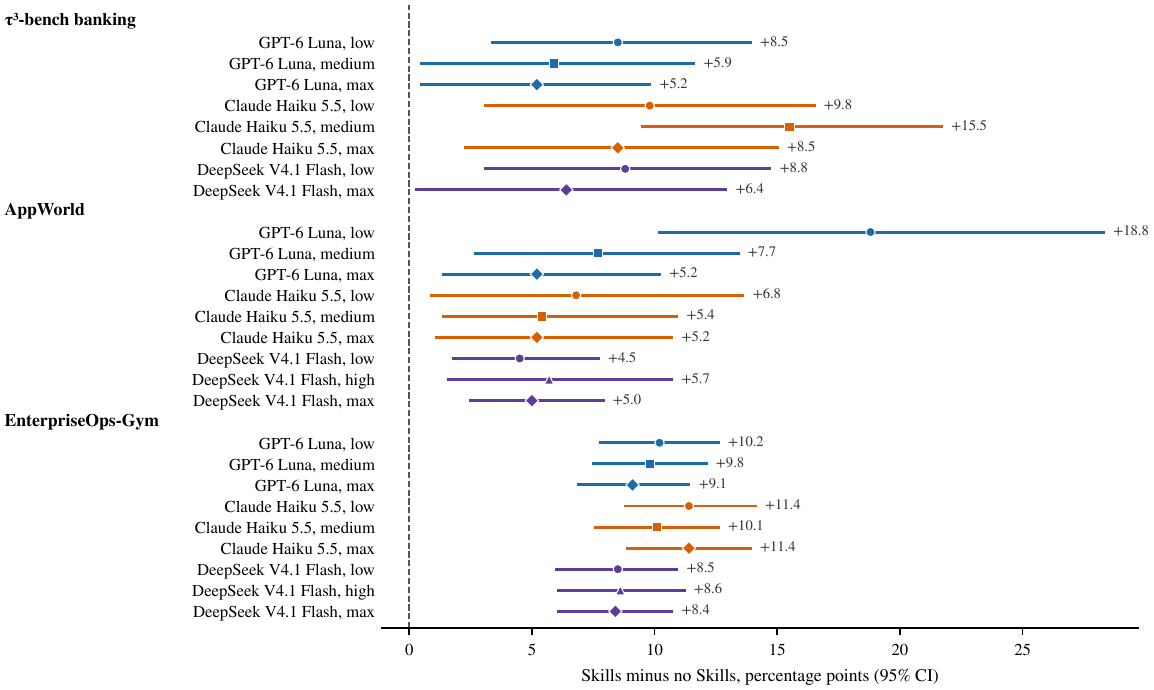}
\caption{Paired difference in pass rate, Skills minus no Skills, for each executor and effort, with 95\% confidence intervals (Tables 8 to 10).}
\end{figure}

\needspace{10\baselineskip}
\subsection{Effort and reliability}\label{effort-and-reliability}

Table 11 compares Skills at the lowest effort with no Skills at the highest effort.

\par\noindent\begin{minipage}{\linewidth}
\needspace{8\baselineskip}
\textbf{Table 11. Skills at the lowest effort against no Skills at the maximum effort.}\par\smallskip\noindent
\begin{tabular}{@{}
  >{\raggedright\arraybackslash}p{\dimexpr 0.2449\linewidth - 1.9592\tabcolsep\relax}
  >{\raggedright\arraybackslash}p{\dimexpr 0.1622\linewidth - 1.2976\tabcolsep\relax}
  >{\raggedright\arraybackslash}p{\dimexpr 0.2177\linewidth - 1.7416\tabcolsep\relax}
  >{\raggedright\arraybackslash}p{\dimexpr 0.2697\linewidth - 2.1576\tabcolsep\relax}
  >{\raggedleft\arraybackslash}p{\dimexpr 0.1055\linewidth - 0.8440\tabcolsep\relax}@{}}
\toprule\noalign{}
\begin{minipage}[b]{\linewidth}\raggedright
Benchmark
\end{minipage} & \begin{minipage}[b]{\linewidth}\raggedright
GPT-6 Luna
\end{minipage} & \begin{minipage}[b]{\linewidth}\raggedright
Claude Haiku 5.5
\end{minipage} & \begin{minipage}[b]{\linewidth}\raggedright
DeepSeek V4.1 Flash
\end{minipage} & \begin{minipage}[b]{\linewidth}\raggedleft
Teacher
\end{minipage} \\
\midrule\noalign{}
\ensuremath{\tau}\ensuremath{^3} & 21.6\% \textless{} 38.9\% & 28.9\% \textless{} 43.3\% & 50.8\% \textgreater{} 46.4\% & 57.2\% \\
AppWorld & 87.3\% \textless{} 91.4\% & 94.8\% \textgreater{} 92.7\% & 97.3\% \textgreater{} 91.8\% & 97.3\% \\
EnterpriseOps-Gym & 37.9\% \textgreater{} 33.3\% & 45.3\% \textgreater{} 35.8\% & 56.8\% \textgreater{} 49.9\% & 52.8\% \\
\bottomrule\noalign{}
\end{tabular}
\end{minipage}\par\medskip

In EnterpriseOps-Gym, the gain from Skills is larger than the gain from the maximum effort for all three models. In \ensuremath{\tau}\ensuremath{^3}, a higher effort gives GPT-6 Luna and Claude Haiku 5.5 more than Skills give. The effort and the Skills add: in each \ensuremath{\tau}\ensuremath{^3} cell, Skills increase the pass rate at the same effort. Figure 5 shows the pass rate at each effort.

\begin{figure}
\centering
\includegraphics[width=1\linewidth,height=\textheight,keepaspectratio]{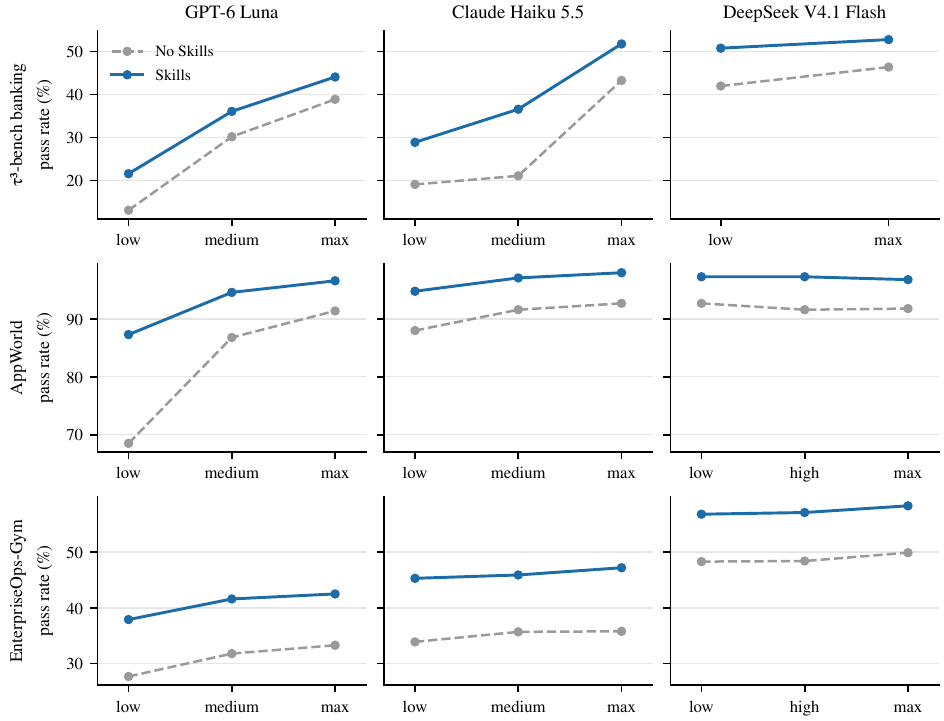}
\caption{Pass rate at each reasoning effort, with and without Skills. Each row is a benchmark, and each column is an executor model.}
\end{figure}

Skills increase the share of tasks that pass in every trial. The \ensuremath{\tau}\ensuremath{^3} pass\ensuremath{^4} count increases in 7 of 8 cells, and it stays at 7 of 97 tasks for Claude Haiku 5.5 \texttt{low} (Figure 6). The AppWorld scenario pass rate increases in each cell, for example from 50.3\% to 72.8\% for GPT-6 Luna \texttt{low} and from 87.1\% to 98.0\% for Claude Haiku 5.5 \texttt{max}.

\begin{figure}
\centering
\includegraphics[width=1\linewidth,height=\textheight,keepaspectratio]{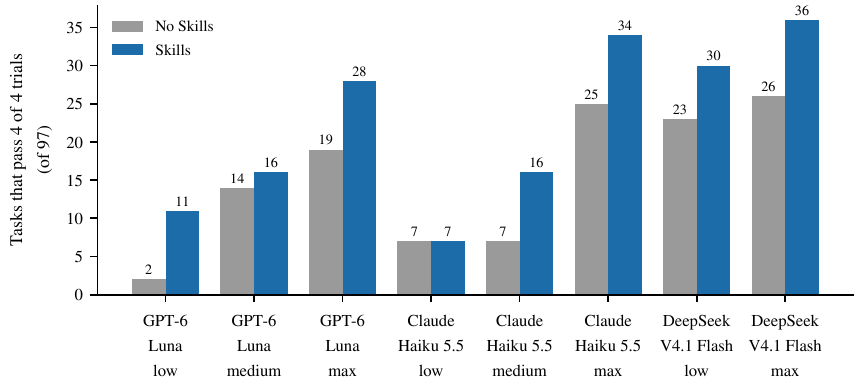}
\caption{\ensuremath{\tau}\ensuremath{^3}-bench banking: tasks that pass all four trials (pass\ensuremath{^4}), of 97 tasks.}
\end{figure}

\needspace{10\baselineskip}
\subsection{Tokens and latency}\label{tokens-and-latency}

The token costs in this section include the cost of the Skill index. The benchmark harness puts the index in the system prompt (Section 6.2), so each model call carries the descriptors of all Skills in its scope. A production agent does not have this index. It calls Skills search, which returns the descriptors of only the matching Skills, and then it reads the body of each Skill that it selects (Section 5.5). The fixed cost of the index is thus a cost of the benchmark harness, not of a production agent.

\par\noindent\begin{minipage}{\linewidth}
\needspace{8\baselineskip}
\textbf{Table 12. Change in executor tokens for each passed episode, and mean latency, Skills against no Skills.} The \ensuremath{\tau}\ensuremath{^3} latency is the mean duration of a passed episode. The AppWorld latency is the mean over all episodes. \ensuremath{\tau}\ensuremath{^3} ran DeepSeek V4.1 Flash only at \texttt{low} and \texttt{max}.\par\smallskip\noindent
\begin{tabular}{@{}
  >{\raggedright\arraybackslash}p{\dimexpr 0.2417\linewidth - 2.4170\tabcolsep\relax}
  >{\raggedright\arraybackslash}p{\dimexpr 0.0830\linewidth - 0.8300\tabcolsep\relax}
  >{\raggedleft\arraybackslash}p{\dimexpr 0.1019\linewidth - 1.0190\tabcolsep\relax}
  >{\raggedleft\arraybackslash}p{\dimexpr 0.1878\linewidth - 1.8780\tabcolsep\relax}
  >{\raggedleft\arraybackslash}p{\dimexpr 0.1498\linewidth - 1.4980\tabcolsep\relax}
  >{\raggedleft\arraybackslash}p{\dimexpr 0.2357\linewidth - 2.3570\tabcolsep\relax}@{}}
\toprule\noalign{}
\begin{minipage}[b]{\linewidth}\raggedright
Executor
\end{minipage} & \begin{minipage}[b]{\linewidth}\raggedright
Effort
\end{minipage} & \begin{minipage}[b]{\linewidth}\raggedleft
\ensuremath{\tau}\ensuremath{^3} tokens
\end{minipage} & \begin{minipage}[b]{\linewidth}\raggedleft
AppWorld tokens
\end{minipage} & \begin{minipage}[b]{\linewidth}\raggedleft
\ensuremath{\tau}\ensuremath{^3} latency (s)
\end{minipage} & \begin{minipage}[b]{\linewidth}\raggedleft
AppWorld latency (s)
\end{minipage} \\
\midrule\noalign{}
GPT-6 Luna & low & \ensuremath{-}7.0\% & +39.0\% & 53 \ensuremath{\rightarrow} 69 & 20.0 \ensuremath{\rightarrow} 24.2 \\
GPT-6 Luna & medium & \ensuremath{-}16.3\% & +42.8\% & 99 \ensuremath{\rightarrow} 95 & 30.0 \ensuremath{\rightarrow} 34.0 \\
GPT-6 Luna & max & \ensuremath{-}28.1\% & +11.5\% & 251 \ensuremath{\rightarrow} 255 & 65.7 \ensuremath{\rightarrow} 55.9 \\
Claude Haiku 5.5 & low & \ensuremath{-}29.2\% & +15.4\% & 41 \ensuremath{\rightarrow} 47 & 14.5 \ensuremath{\rightarrow} 11.8 \\
Claude Haiku 5.5 & medium & \ensuremath{-}44.6\% & +24.0\% & 49 \ensuremath{\rightarrow} 50 & 15.5 \ensuremath{\rightarrow} 13.2 \\
Claude Haiku 5.5 & max & \ensuremath{-}25.0\% & \ensuremath{-}7.9\% & 272 \ensuremath{\rightarrow} 256 & 51.4 \ensuremath{\rightarrow} 37.8 \\
DeepSeek V4.1 Flash & low & \ensuremath{-}20.6\% & +43.5\% & 83 \ensuremath{\rightarrow} 78 & 48.5 \ensuremath{\rightarrow} 43.2 \\
DeepSeek V4.1 Flash & high & --- & +28.9\% & --- & 60.6 \ensuremath{\rightarrow} 49.3 \\
DeepSeek V4.1 Flash & max & \ensuremath{-}23.9\% & \ensuremath{-}22.6\% & 87 \ensuremath{\rightarrow} 83 & 103.7 \ensuremath{\rightarrow} 65.5 \\
\bottomrule\noalign{}
\end{tabular}
\end{minipage}\par\medskip

Figure 7 shows the token values of Table 12.

\begin{figure}
\centering
\includegraphics[width=1\linewidth,height=\textheight,keepaspectratio]{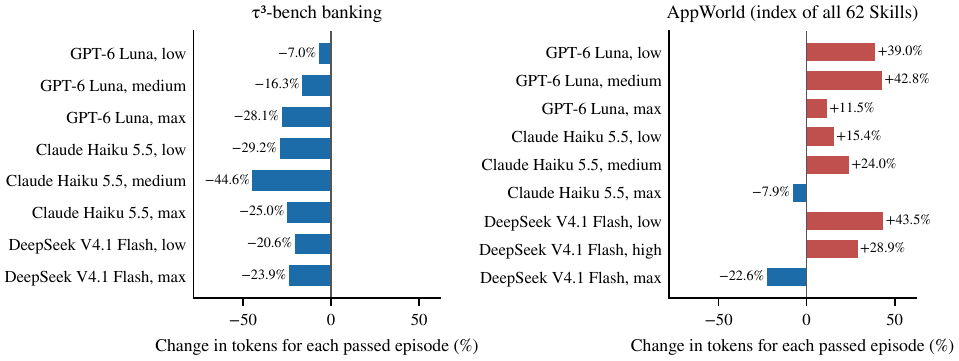}
\caption{Change in executor tokens for each passed episode, Skills against no Skills (Table 12). The AppWorld harness gave the agent an index of all 62 Skills.}
\end{figure}

\textbf{\ensuremath{\tau}\ensuremath{^3}.} Skills decrease the tokens for each passed episode in all eight cells. A passed \ensuremath{\tau}\ensuremath{^3} episode uses 0.2 to 2.3 million tokens, so the index is a small part of the prompt. The total tokens of a cell decrease in five of eight cells (\ensuremath{-}4.0\% to \ensuremath{-}18.6\%) and increase most for GPT-6 Luna \texttt{low} (+53.2\%). The mean duration of a passed episode decreases in four of eight cells. It increases most at \texttt{low} effort for GPT-6 Luna (53 s to 69 s) and Claude Haiku 5.5 (41 s to 47 s). The duration includes the time to record each episode as a Trace.

\textbf{AppWorld.} The tokens for each passed episode increase in seven of nine cells. An AppWorld episode is short (about 55,000 to 260,000 prompt tokens), and the index of all 62 Skills added about 2,300 prompt tokens to each model call. A family index would add fewer tokens. Most prompt tokens are cached in both arms. The tokens decrease in the two cells with the longest episodes, Claude Haiku 5.5 \texttt{max} and DeepSeek V4.1 Flash \texttt{max}. Skills decrease the steps and the latency of Claude Haiku 5.5 and DeepSeek V4.1 Flash, for example 14.5 steps against 18.4 for Claude Haiku 5.5 \texttt{max}. GPT-6 Luna at \texttt{low} and \texttt{medium} is slower with Skills.

\textbf{EnterpriseOps-Gym.} The \texttt{read\_skill} calls add 0.7 to 2.3 model calls to each episode. The passed episodes of GPT-6 Luna, and of Claude Haiku 5.5 at \texttt{low} and \texttt{medium}, are 1 to 4 seconds slower at the median. The passed episodes of Claude Haiku 5.5 \texttt{max} are faster (59 s against 68 s at the median).

These results agree with a fixed cost that long tasks amortize: the index adds tokens to each model call, and a Skill can save steps. No arm isolates the cost of the index, so this explanation is an interpretation. The latency results have many confounds: the Skills arm makes more model calls, each duration includes the time to record the Trace, the latency of each model API varies, for the third-party APIs and for our own model server, and each latency includes the network path from the harness to the model API (Section 6.4). This paper does not draw a conclusion about latency.

\needspace{10\baselineskip}
\subsection{Where the gain comes from}\label{where-the-gain-comes-from}

The teacher result of each task divides the tasks into strata.

\begin{itemize}
\tightlist
\item
  \textbf{\ensuremath{\tau}\ensuremath{^3}} (pooled over eight cells): tasks that the teacher passed in four of four trials gain +17.0 points {[}+11.6, +22.3{]} (40 tasks). Tasks with one to three passes gain +5.0 {[}\ensuremath{-}0.5, +10.5{]} (27 tasks). Tasks that the teacher never passed gain +0.6 {[}\ensuremath{-}2.2, +4.0{]} (30 tasks).
\item
  \textbf{EnterpriseOps-Gym:} tasks that the teacher passed gain +11.0 to +18.2 points (330 tasks). Tasks that the teacher failed gain +2.0 to +5.6 points (295 tasks), and eight of nine intervals are above 0. These tasks have no successful Trace of their own, so this gain is some indication of transfer.
\item
  \textbf{AppWorld:} the 96 tasks whose own teacher Trace a Skill cites gain +5.9 to +22.2 points. The 48 other tasks in the same scenarios have a Skill for their scenario, but no Skill cites their own Trace. These tasks gain \ensuremath{-}1.4 to +2.1 points, except +9.0 for GPT-6 Luna \texttt{low}.
\end{itemize}

Figure 8 shows the \ensuremath{\tau}\ensuremath{^3} strata.

\begin{figure}
\centering
\includegraphics[width=0.65\linewidth,height=\textheight,keepaspectratio]{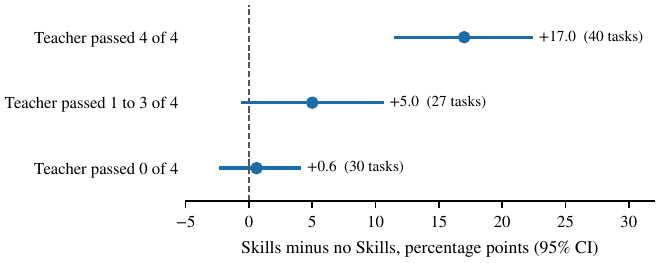}
\caption{\ensuremath{\tau}\ensuremath{^3}-bench banking: gain by teacher result, pooled over eight cells.}
\end{figure}

The gain is largest on the tasks that the teacher solved. This result agrees with the design, because a Skill carries a procedure from a strongly verified success (Section 5.3). In \ensuremath{\tau}\ensuremath{^3}, no Skill cites a Trace of a task that the teacher never passed, and the gain on these tasks is near 0. In EnterpriseOps-Gym, the gain on the teacher-failed tasks comes from the Skills of other tasks or from the other differences of the Skills arm (Section 6.2).

In EnterpriseOps-Gym, the episodes that read the Skill of their own task pass 62.6\% to 83.3\% of the time. The episodes that read only other Skills pass 28.3\% to 47.2\% of the time. This comparison is descriptive. The episodes with their own Skill are mostly on tasks that the teacher passed, and these tasks are easier.

\textbf{Transfer.} The main experiments do not test transfer (Section 6.1). However, four results give some evidence that transfer occurs:

\begin{itemize}
\tightlist
\item
  \textbf{EnterpriseOps-Gym tasks that the teacher failed} gain in eight of nine cells, with intervals above 0 (see above). The experiment does not separate the effect of the Skills of other tasks from the other differences of the Skills arm.
\item
  \textbf{ALFWorld} tests transfer directly. Zep compiled the Skills from the runs of the teacher on train games. The experiment scores the \texttt{valid\_unseen} games, which are in rooms that no train game uses. Thus, no Skill comes from a scored game or from its room, and the gain on these games (Section 7.6) comes from the Skills of other games.
\item
  \textbf{AppWorld scenario \texttt{b0a8eae}} has no Skill of its own, because the teacher failed all three variants. GPT-6 Luna \texttt{low} passed 6 of 9 episodes with a Skill from another scenario and 0 of 9 without Skills. This result comes from one scenario only.
\item
  \textbf{An earlier EnterpriseOps-Gym split experiment} compiled Skills from one part of the tasks and scored 152 other tasks. Four cells, all at the \texttt{xhigh} effort of that run, had intervals above 0 (+3.9 to +5.3 points). The intervals of the other cells include 0.
\end{itemize}

The main experiments were not designed to test transfer, so these results do not conclusively establish that transfer occurs (Section 9).

\needspace{10\baselineskip}
\subsection{Comparison with continuous fine-tuning (ALFWorld)}\label{comparison-with-continuous-fine-tuning-alfworld}

RSD \citep{rsd} trains the weights of the model in several rounds of on-policy fine-tuning. Zep Agent Skills changes only the context of the agent. Both methods learn from the runs of the same model.

\par\noindent\begin{minipage}{\linewidth}
\needspace{8\baselineskip}
\textbf{Table 13. ALFWorld \texttt{valid\_unseen}.} The CI is a 95\% paired bootstrap interval over games. The teacher is the executor model (Qwen3-8B) on train games, so the table has no separate teacher row. ``Failures removed'' is the gain divided by the failure rate of the base.\par\smallskip\noindent
\begin{tabular}{@{}
  >{\raggedright\arraybackslash}p{\dimexpr 0.2358\linewidth - 2.8296\tabcolsep\relax}
  >{\raggedright\arraybackslash}p{\dimexpr 0.1247\linewidth - 1.4964\tabcolsep\relax}
  >{\raggedright\arraybackslash}p{\dimexpr 0.1685\linewidth - 2.0220\tabcolsep\relax}
  >{\raggedleft\arraybackslash}p{\dimexpr 0.0648\linewidth - 0.7776\tabcolsep\relax}
  >{\raggedleft\arraybackslash}p{\dimexpr 0.0822\linewidth - 0.9864\tabcolsep\relax}
  >{\raggedleft\arraybackslash}p{\dimexpr 0.2190\linewidth - 2.6280\tabcolsep\relax}
  >{\raggedleft\arraybackslash}p{\dimexpr 0.1050\linewidth - 1.2600\tabcolsep\relax}@{}}
\toprule\noalign{}
\begin{minipage}[b]{\linewidth}\raggedright
Method
\end{minipage} & \begin{minipage}[b]{\linewidth}\raggedright
Model
\end{minipage} & \begin{minipage}[b]{\linewidth}\raggedright
Training
\end{minipage} & \begin{minipage}[b]{\linewidth}\raggedleft
Base
\end{minipage} & \begin{minipage}[b]{\linewidth}\raggedleft
Result
\end{minipage} & \begin{minipage}[b]{\linewidth}\raggedleft
Gain, points
\end{minipage} & \begin{minipage}[b]{\linewidth}\raggedleft
Failures removed
\end{minipage} \\
\midrule\noalign{}
Zep Agent Skills, weak base (\texttt{fixed}) & Qwen3-8B & None & 42.8\% & 75.4\% & +32.6 {[}+24.9, +40.3{]} & 57\% \\
RSD (paper) & Qwen3-8B & Several rounds & 62.9\% & 76.6\% & +13.7 & 37\% \\
RSD (paper) & Qwen3-14B & Several rounds & 69.3\% & 79.6\% & +10.3 & 34\% \\
Zep Agent Skills, strong base (\texttt{typed}) & Qwen3-8B & None & 74.6\% & 82.8\% & +8.2 {[}+2.5, +13.9{]} & 32\% \\
Harness@Inf (paper, context only) & Qwen3-8B & None & 62.9\% & 65.4\% & +2.5 & 7\% \\
\bottomrule\noalign{}
\end{tabular}
\end{minipage}\par\medskip

Figure 9 shows the bases and the results of Table 13.

\begin{figure}
\centering
\includegraphics[width=1\linewidth,height=\textheight,keepaspectratio]{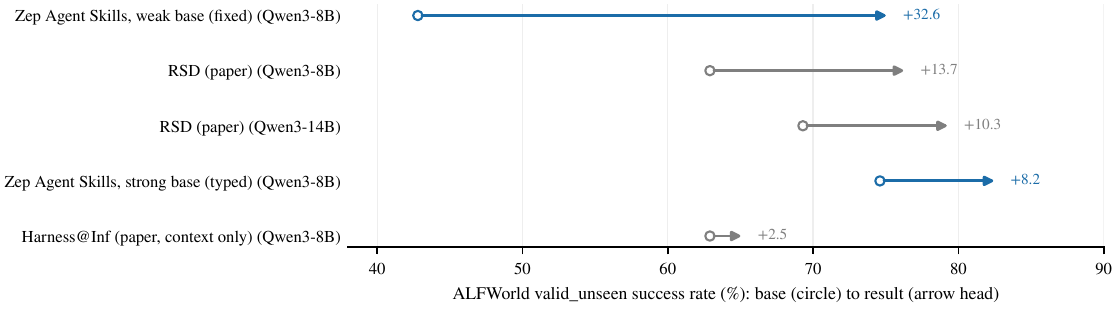}
\caption{ALFWorld valid\_unseen: base and result of each method. The RSD rows are the values that the RSD paper reports. The bases and the harnesses differ, so the figure compares gains, not absolute results.}
\end{figure}

From the weak base, Zep Agent Skills removes 57\% of the failures. RSD removes 37\%. From the strong base, the Zep Agent Skills result (82.8\%) is higher than RSD on Qwen3-8B (76.6\%) and on Qwen3-14B (79.6\%). The strong-base gain (+8.2) is less than the RSD gain (+13.7), and its interval includes the RSD gain. The context-only baseline of the RSD paper (Harness@Inf) adds 2.5 points.

The gain from the weak base comes from the procedures that the base does not know. The heat task type increases from 4\% to 65\%, the clean type from 38\% to 90\%, and the cool type from 29\% to 63\%. The ``two objects'' type does not increase (41\% to 37\%). This type is the weakest type in all setups.

On \texttt{valid\_seen} (games in rooms that the train split contains), the weak-base gain is +16.4 points {[}+9.5, +23.3{]}, and the strong-base gain is +1.7 {[}\ensuremath{-}4.5, +7.9{]}.

\needspace{10\baselineskip}
\subsection{Skill use, failure types, and losses}\label{skill-use-failure-types-and-losses}

The executors read a Skill in most episodes (Table 14).

\par\noindent\begin{minipage}{\linewidth}
\needspace{8\baselineskip}
\textbf{Table 14. Episodes with a \texttt{read\_skill} call.}\par\smallskip\noindent
\begin{tabular}{@{}
  >{\raggedright\arraybackslash}p{\dimexpr 0.3701\linewidth - 0.7402\tabcolsep\relax}
  >{\raggedright\arraybackslash}p{\dimexpr 0.6299\linewidth - 1.2598\tabcolsep\relax}@{}}
\toprule\noalign{}
\begin{minipage}[b]{\linewidth}\raggedright
Benchmark
\end{minipage} & \begin{minipage}[b]{\linewidth}\raggedright
Episodes with a \texttt{read\_skill} call
\end{minipage} \\
\midrule\noalign{}
\ensuremath{\tau}\ensuremath{^3} & 93.0\% to 100\% \\
AppWorld & 97.7\% to 100\% \\
EnterpriseOps-Gym & 93.3\% to 100\% \\
\bottomrule\noalign{}
\end{tabular}
\end{minipage}\par\medskip

In EnterpriseOps-Gym, Claude Haiku 5.5 at \texttt{low} and \texttt{medium} did not read a Skill in 4\% to 7\% of the episodes. These episodes passed 16\% to 18\% of the time. Because the current executors read a Skill in most episodes, the current runs cannot show whether a read is necessary for a gain.

\textbf{Failure types.} In EnterpriseOps-Gym, the Skills arm has fewer episodes that end in a wrong final state in all nine cells (for example, 930 against 1,052 for GPT-6 Luna \texttt{low}). The Skills arm also has fewer episodes that stop for missing information in all nine cells. For DeepSeek V4.1 Flash, the Skills arm has more episodes that reach the step limit (9 to 17, against 5 to 8). A rule-based classifier assigns these categories.

\textbf{Losses.} For each EnterpriseOps-Gym cell, Skills won 73 to 94 tasks and lost 14 to 32 tasks. A task is won when the Skills arm passes two or more of three trials and the no-Skills arm passes one or fewer. In 3 to 9 of the lost tasks of each cell, the agent read the Skill of the same task and the teacher passed the task.

One loss shows how a Skill can cause a failure. In the AppWorld scenario \texttt{229360a}, two variants use opposite rules (``liked and downloaded'' against ``liked or downloaded''). The compiler kept one rule, and Skills lose on this scenario for two cells.

\needspace{10\baselineskip}
\section{Analysis}\label{analysis}

Zep Agent Skills is designed so that an agent does not solve a task again after another agent solved it and a verifier confirmed the result. The results in Section 7 support this design on the tasks that the teacher solved, and they show the cases where Skills do not help.

Zep Agent Skills helps in five ways.

\textbf{A weaker model repeats the work of a stronger model.} Skills increase the pass rate in all 26 cells (Section 7.2), and each executor moves toward the teacher. For example, Claude Haiku 5.5 at \texttt{medium} effort on \ensuremath{\tau}\ensuremath{^3} goes from 21.1\% to 36.6\%, against 57.2\% for the teacher. On EnterpriseOps-Gym, DeepSeek V4.1 Flash with Skills passes more tasks than the single teacher run (56.8\% to 58.3\%, against 52.8\%). This is the production pattern of Section 6.2: a stronger model solves a task one time, and a cheaper model then does the task again with the Skill. On EnterpriseOps-Gym, Skills at the lowest effort also pass more tasks than no Skills at the maximum effort, for all three executors (Section 7.3).

\textbf{The gain comes from the verified procedure.} The Skills arm also has a longer prompt and one more tool. These differences are the same for each task, but the gain is largest on the tasks whose own successful Trace a Skill cites (Section 7.5):

\begin{itemize}
\tightlist
\item
  \textbf{\ensuremath{\tau}\ensuremath{^3}:} +17.0 points on the tasks that the teacher passed in all four trials, against +0.6 on the tasks that it never passed.
\item
  \textbf{AppWorld:} +5.9 to +22.2 points on the tasks whose Trace a Skill cites, against \ensuremath{-}1.4 to +2.1 on the other tasks of the same scenarios, except +9.0 for GPT-6 Luna \texttt{low}.
\end{itemize}

Thus, the content of the Skill is the most probable cause of most of the gain. The failure types agree with this result. In EnterpriseOps-Gym, the Skills arm has fewer episodes that end in a wrong final state, and fewer episodes that stop for missing information, in all nine cells (Section 7.7). A Skill gives the executor the steps and the information sources that the teacher used.

\textbf{The result is more repeatable.} Skills increase the \ensuremath{\tau}\ensuremath{^3} tasks that pass in all four trials in 7 of 8 cells, and the AppWorld scenario pass rate in 9 of 9 cells (Section 7.3).

\textbf{Long tasks use fewer tokens.} On \ensuremath{\tau}\ensuremath{^3}, Skills decrease the tokens for each passed episode in all eight cells, by 7.0\% to 44.6\% (Section 7.4). On these long tasks, the steps that a Skill saves cost more than the Skill index adds.

\textbf{A model learns new procedures without training.} On ALFWorld, the Skills from the runs of Qwen3-8B on train games increase its success on games in new rooms by 32.6 points from the weak base. They remove 57\% of the failures, and RSD removes 37\% (Section 7.6). From the strong base, they remove 32\% of the failures. The gain is on the task types that the base did not know: for example, the heat type increases from 4\% to 65\%. The Skills come from other games, so this result is also the clearest evidence of transfer. The EnterpriseOps-Gym tasks that the teacher failed give weaker evidence: they gain +2.0 to +5.6 points (Section 7.5).

Zep Agent Skills does not help in four cases.

\textbf{Tasks that the teacher did not solve.} The \ensuremath{\tau}\ensuremath{^3} tasks that the teacher never passed gain only +0.6 points {[}\ensuremath{-}2.2, +4.0{]} (Section 7.5). The compiler uses only strongly verified successes, so these tasks get no Skill of their own. The ability of the teacher thus limits the gain. See Section 9 for a future research direction.

\textbf{Tasks that the agent already knows how to do.} When the base agent already has the procedure, a Skill adds little. AppWorld is saturated: the no-Skills arm passes 68.5\% to 92.7\%, and in eight of nine cells the gain is 4.5 to 7.7 points (Section 7.2). On ALFWorld, the gain from the strong base is +8.2 points on \texttt{valid\_unseen} and +1.7 {[}\ensuremath{-}4.5, +7.9{]} on \texttt{valid\_seen} (Section 7.6).

\textbf{Losses from the Skill text.} In each EnterpriseOps-Gym cell, Skills lose 14 to 32 tasks, against 73 to 94 wins (Section 7.7). The observed loss mechanism is a risk of the compiler design: a merge of two procedures that are similar but have opposite rules can keep the incorrect rule for one task (Section 5.3). This risk is an area of ongoing work.

\textbf{Short tasks with a full Skill index.} On AppWorld, the tokens for each passed episode increase in seven of nine cells, because the index of all 62 Skills adds about 2,300 prompt tokens to each model call (Section 7.4). A production agent uses Skills search, not this index.

\needspace{10\baselineskip}
\section{Limitations and future work}\label{limitations-and-future-work}

\textbf{Transfer.} The main experiments score the tasks that the teacher ran, and most of the gain comes from the tasks that the teacher succeeded on. They show that compiled Skills help an agent repeat a known task, and they do not show a gain on new tasks. The benchmarks limit this test: most of their tasks share little similarity with the other tasks (Section 6.1). A transfer test needs tasks that share procedures but differ in their instances.

\textbf{Learning only from successes.} The compiler uses only strongly verified successes as support (Section 5.3). A task that the teacher never solved thus gets no Skill of its own, and its gain is small or 0 (Section 7.5). A failed Trace can still contain parts that succeeded, for example a sub-goal that a verifier checked before a later step failed. Future work will compile Skills from these verified parts of failed Traces. The sub-goals can come from checks that a verifier records during the task, or they can be inferred from the Traces of the task family.

\needspace{10\baselineskip}
\section{Conclusion}\label{conclusion}

Memory systems for production LLM agents mostly store facts and earlier interactions, not procedures that a verifier has tested. Zep Agent Skills captures each task attempt as a Trace, compiles strongly verified successes into versioned Skills that cite their evidence, curates them against the current library, admits them through trusted checks, and serves them in the open Agent Skills format. Automatic controls address the risks of a service that can admit a Skill without human review: instance-data leakage, instruction injection, poisoned outcomes, and flow across boundaries. On \ensuremath{\tau}\ensuremath{^3}-bench, AppWorld, and EnterpriseOps-Gym, the compiled Skills increase the pass rate of each executor model at each reasoning effort, by 4.5 to 18.8 points. The gain is largest on the tasks that the teacher solved, which agrees with a design that learns from verified successes, so the success of the teacher limits the gain. The Skill index of the benchmark harness adds a cost to each model call, and Skills cause a small number of losses. On ALFWorld, Skills from the runs of a small model give a gain similar to or larger than the gain that a fine-tuning method reports. Two questions remain open: transfer to new tasks in the main benchmarks, and learning from tasks that the agent fails.

\bibliographystyle{unsrtnat}
\bibliography{references}

@article{coala,
  title = {{Cognitive Architectures for Language Agents}},
  author = {Sumers, Theodore R. and Yao, Shunyu and Narasimhan, Karthik and Griffiths, Thomas L.},
  journal = {arXiv preprint arXiv:2309.02427},
  year = {2023},
  eprint = {2309.02427},
  archivePrefix = {arXiv},
  url = {https://arxiv.org/abs/2309.02427}
}

@article{reflexion,
  title = {{Reflexion: Language Agents with Verbal Reinforcement Learning}},
  author = {Shinn, Noah and Cassano, Federico and Berman, Edward and Gopinath, Ashwin and Narasimhan, Karthik and Yao, Shunyu},
  journal = {arXiv preprint arXiv:2303.11366},
  year = {2023},
  eprint = {2303.11366},
  archivePrefix = {arXiv},
  url = {https://arxiv.org/abs/2303.11366}
}

@article{voyager,
  title = {{Voyager: An Open-Ended Embodied Agent with Large Language Models}},
  author = {Wang, Guanzhi and Xie, Yuqi and Jiang, Yunfan and Mandlekar, Ajay and Xiao, Chaowei and Zhu, Yuke and Fan, Linxi and Anandkumar, Anima},
  journal = {arXiv preprint arXiv:2305.16291},
  year = {2023},
  eprint = {2305.16291},
  archivePrefix = {arXiv},
  url = {https://arxiv.org/abs/2305.16291}
}

@article{expel,
  title = {{ExpeL: LLM Agents Are Experiential Learners}},
  author = {Zhao, Andrew and Huang, Daniel and Xu, Quentin and Lin, Matthieu and Liu, Yong-Jin and Huang, Gao},
  journal = {arXiv preprint arXiv:2308.10144},
  year = {2023},
  eprint = {2308.10144},
  archivePrefix = {arXiv},
  url = {https://arxiv.org/abs/2308.10144}
}

@article{awm,
  title = {{Agent Workflow Memory}},
  author = {Wang, Zora Zhiruo and Mao, Jiayuan and Fried, Daniel and Neubig, Graham},
  journal = {arXiv preprint arXiv:2409.07429},
  year = {2024},
  eprint = {2409.07429},
  archivePrefix = {arXiv},
  url = {https://arxiv.org/abs/2409.07429}
}

@article{memp,
  title = {{Memp: Exploring Agent Procedural Memory}},
  author = {Fang, Runnan and Liang, Yuan and Wang, Xiaobin and Wu, Jialong and Qiao, Shuofei and Xie, Pengjun and Huang, Fei and Chen, Huajun and Zhang, Ningyu},
  journal = {arXiv preprint arXiv:2508.06433},
  year = {2025},
  eprint = {2508.06433},
  archivePrefix = {arXiv},
  url = {https://arxiv.org/abs/2508.06433}
}

@article{ace,
  title = {{Agentic Context Engineering: Evolving Contexts for Self-Improving Language Models}},
  author = {Zhang, Qizheng and Hu, Changran and Upasani, Shubhangi and Ma, Boyuan and Hong, Fenglu and Kamanuru, Vamsidhar and Rainton, Jay and Wu, Chen and Ji, Mengmeng and Li, Hanchen and Thakker, Urmish and Zou, James and Olukotun, Kunle},
  journal = {arXiv preprint arXiv:2510.04618},
  year = {2025},
  eprint = {2510.04618},
  archivePrefix = {arXiv},
  url = {https://arxiv.org/abs/2510.04618}
}

@article{legomem,
  title = {{LEGOMem: Modular Procedural Memory for Multi-agent LLM Systems for Workflow Automation}},
  author = {Han, Dongge and Couturier, Camille and Diaz, Daniel Madrigal and Zhang, Xuchao and Rühle, Victor and Rajmohan, Saravan},
  journal = {arXiv preprint arXiv:2510.04851},
  year = {2025},
  eprint = {2510.04851},
  archivePrefix = {arXiv},
  url = {https://arxiv.org/abs/2510.04851}
}

@article{skillpro,
  title = {{Skill-Pro: Learning Reusable Skills from Experience via Non-Parametric PPO for LLM Agents}},
  author = {Mi, Qirui and Ma, Zhijian and Yang, Mengyue and Li, Haoxuan and Wang, Yisen and Zhang, Haifeng and Wang, Jun},
  journal = {arXiv preprint arXiv:2602.01869},
  year = {2026},
  eprint = {2602.01869},
  archivePrefix = {arXiv},
  url = {https://arxiv.org/abs/2602.01869}
}

@article{skillrevise,
  title = {{SkillRevise: Improving LLM-Authored Agent Skills via Trace-Conditioned Skill Revision}},
  author = {Liu, Yuxuan and Su, Zhaochen and Xie, Lingyun and Zhang, Yuhao and Zong, Qing and Guo, Jiahe and Xie, Zhongwei and Ji, Yiyan and Yim, Yauwai and Luo, Hongyu and Ren, Xiyu and Chenyu, Ruan and Li, Haoran and Song, Yangqiu},
  journal = {arXiv preprint arXiv:2606.01139},
  year = {2026},
  eprint = {2606.01139},
  archivePrefix = {arXiv},
  url = {https://arxiv.org/abs/2606.01139}
}

@article{anything2skill,
  title = {{Anything2Skill: Compiling External Knowledge into Reusable Skills for Agents}},
  author = {Pan, Qianjun and Yang, Yutao and Li, Junsong and Zhou, Jie and Chen, Kai and Li, Xin and Chen, Qin and He, Liang},
  journal = {arXiv preprint arXiv:2606.09316},
  year = {2026},
  eprint = {2606.09316},
  archivePrefix = {arXiv},
  url = {https://arxiv.org/abs/2606.09316}
}

@article{wikiskill,
  title = {{WikiSkill: Compiling Agent Experience into Persistent Knowledge for Skill Evolution}},
  author = {Tang, Liyan and Rashtchian, Cyrus and Ferng, Chun-Sung and Tomkins, Andrew and Juan, Da-Cheng and Vu, Tu},
  journal = {arXiv preprint arXiv:2608.27454},
  year = {2026},
  eprint = {2608.27454},
  archivePrefix = {arXiv},
  url = {https://arxiv.org/abs/2608.27454}
}

@article{proceduralgraphs,
  title = {{Procedural Graphs: Self-Evolving Execution Structures for LLM Agents}},
  author = {Lu, Yuxing and Chen, Yicheng and Wu, Shanchan and Arık, Sercan Ö.},
  journal = {arXiv preprint arXiv:2609.09153},
  year = {2026},
  eprint = {2609.09153},
  archivePrefix = {arXiv},
  url = {https://arxiv.org/abs/2609.09153}
}

@article{rsd,
  title = {{Self-Evolving Agents through Privileged Retrospective Distillation}},
  author = {Zhou, Yuhang and Wu, Yifan and Peng, Bo and Wang, Mingyi and Liu, Jiayi and Fan, Xiangjun and Zhang, Lizhu and Zhao, Zhuokai},
  journal = {arXiv preprint arXiv:2610.27495},
  year = {2026},
  eprint = {2610.27495},
  archivePrefix = {arXiv},
  url = {https://arxiv.org/abs/2610.27495}
}

@article{skillsbench,
  title = {{SkillsBench: Benchmarking How Well Agent Skills Work Across Diverse Tasks}},
  author = {Li, Xiangyi and Liu, Yimin and Chen, Wenbo and You, Bingran and Di, Zonglin and He, Yifeng and Zheng, Shenghan and Choe, Kyoung Whan and Sun, Jiankai and Wang, Shuyi and Tao, Chujun and Li, Binxu and Zhao, Xuandong and Geng, Hejia and Wu, Xiaojun and Zhou, Junwei and Chen, Xiaokun and Xing, Hanwen and Li, Yubo and Zeng, Qunhong and Wang, Di and Wang, Yuanli and Chaim, Roey Ben and Jiang, Penghao and Shen, Haotian and Kong, Luyang and Liu, Xinyi and Wang, Runhui and Liu, Xuanqing and Li, Jiachen and Lan, Xin and Lin, Yueqian and Ye, Wengao and He, Junwei and Li, Songlin and Zhang, Yue and Gao, Yipeng and Li, Yijiang and Ma, Ze and Jing, Liqiang and Wang, Tianyu and Li, Kaixin and Xue, Yiqi and Lyu, Haoran and He, Yizhuo and Tian, Yuchen and Wu, Shutong and Wang, Bowei and Gao, Yixuan and Chen, Bo and Liu, Litong and Cheng, Sikai and Bao, Jiajun and Tong, Shuaicheng and Xu, Shuwen and Zhuo, Terry Yue and Ye, Tinghan and Qi, Qi and Li, Miao and Liao, Longtai and Tan, Zelin and Shi, Chang and Tang, Xilin and Tankasala, Srinath and Yuan, Boqin and Qian, Yaoyao and Tu, Jianhong and Wang, Chenguang and Sun, Yizhou and Wang, Wei and Taylor, Aaron and Yang, Ziyue and Guan, Changkun and Dong, Zhikang and Zhang, Xinyu and Dillmann, Steven and Lee, Han-chung and Song, Dawn},
  journal = {arXiv preprint arXiv:2602.12670},
  year = {2026},
  eprint = {2602.12670},
  archivePrefix = {arXiv},
  url = {https://arxiv.org/abs/2602.12670}
}

@article{autoskill,
  title = {{AutoSkill: Experience-Driven Lifelong Learning via Skill Self-Evolution}},
  author = {Yang, Yutao and Li, Junsong and Pan, Qianjun and Zhan, Bihao and Cai, Yuxuan and Du, Lin and Zhou, Jie and Chen, Kai and Chen, Qin and Li, Xin and Zhang, Bo and He, Liang},
  journal = {arXiv preprint arXiv:2603.01145},
  year = {2026},
  eprint = {2603.01145},
  archivePrefix = {arXiv},
  url = {https://arxiv.org/abs/2603.01145}
}

@article{trace2skill,
  title = {{Trace2Skill: Distill Trajectory-Local Lessons into Transferable Agent Skills}},
  author = {Ni, Jingwei and Liu, Yihao and Liu, Xinpeng and Sun, Yutao and Zhou, Mengyu and Cheng, Pengyu and Wang, Dexin and Zhao, Erchao and Jiang, Xiaoxi and Jiang, Guanjun},
  journal = {arXiv preprint arXiv:2603.25158},
  year = {2026},
  eprint = {2603.25158},
  archivePrefix = {arXiv},
  url = {https://arxiv.org/abs/2603.25158}
}

@article{dynamicskills,
  title = {{Dynamic Agent Skills: A Lifecycle Survey and Taxonomy of Evolving Skill Libraries}},
  author = {Li, Yubo},
  journal = {arXiv preprint arXiv:2607.10113},
  year = {2026},
  eprint = {2607.10113},
  archivePrefix = {arXiv},
  url = {https://arxiv.org/abs/2607.10113}
}

@article{skilllift,
  title = {{SkillLift: Learning Dense Rubrics from Sparse Oracles for Efficient Skill Evolution}},
  author = {Kang, Haoxiang and Wen, Ming},
  journal = {arXiv preprint arXiv:2609.15396},
  year = {2026},
  eprint = {2609.15396},
  archivePrefix = {arXiv},
  url = {https://arxiv.org/abs/2609.15396}
}

@article{memgpt,
  title = {{MemGPT: Towards LLMs as Operating Systems}},
  author = {Packer, Charles and Wooders, Sarah and Lin, Kevin and Fang, Vivian and Patil, Shishir G. and Stoica, Ion and Gonzalez, Joseph E.},
  journal = {arXiv preprint arXiv:2310.08560},
  year = {2023},
  eprint = {2310.08560},
  archivePrefix = {arXiv},
  url = {https://arxiv.org/abs/2310.08560}
}

@article{zep,
  title = {{Zep: A Temporal Knowledge Graph Architecture for Agent Memory}},
  author = {Rasmussen, Preston and Paliychuk, Pavlo and Beauvais, Travis and Ryan, Jack and Chalef, Daniel},
  journal = {arXiv preprint arXiv:2501.13956},
  year = {2025},
  eprint = {2501.13956},
  archivePrefix = {arXiv},
  url = {https://arxiv.org/abs/2501.13956}
}

@article{taubench,
  title = {{$\tau$-bench: A Benchmark for Tool-Agent-User Interaction in Real-World Domains}},
  author = {Yao, Shunyu and Shinn, Noah and Razavi, Pedram and Narasimhan, Karthik},
  journal = {arXiv preprint arXiv:2406.12045},
  year = {2024},
  eprint = {2406.12045},
  archivePrefix = {arXiv},
  url = {https://arxiv.org/abs/2406.12045}
}

@article{tau2,
  title = {{$\tau^2$-Bench: Evaluating Conversational Agents in a Dual-Control Environment}},
  author = {Barres, Victor and Dong, Honghua and Ray, Soham and Si, Xujie and Narasimhan, Karthik},
  journal = {arXiv preprint arXiv:2506.07982},
  year = {2025},
  eprint = {2506.07982},
  archivePrefix = {arXiv},
  url = {https://arxiv.org/abs/2506.07982}
}

@article{tauknowledge,
  title = {{$\tau$-Knowledge: Evaluating Conversational Agents over Unstructured Knowledge}},
  author = {Shi, Quan and Zytek, Alexandra and Razavi, Pedram and Narasimhan, Karthik and Barres, Victor},
  journal = {arXiv preprint arXiv:2603.04370},
  year = {2026},
  eprint = {2603.04370},
  archivePrefix = {arXiv},
  url = {https://arxiv.org/abs/2603.04370}
}

@article{appworld,
  title = {{AppWorld: A Controllable World of Apps and People for Benchmarking Interactive Coding Agents}},
  author = {Trivedi, Harsh and Khot, Tushar and Hartmann, Mareike and Manku, Ruskin and Dong, Vinty and Li, Edward and Gupta, Shashank and Sabharwal, Ashish and Balasubramanian, Niranjan},
  journal = {arXiv preprint arXiv:2407.18901},
  year = {2024},
  eprint = {2407.18901},
  archivePrefix = {arXiv},
  url = {https://arxiv.org/abs/2407.18901}
}

@article{alfworld,
  title = {{ALFWorld: Aligning Text and Embodied Environments for Interactive Learning}},
  author = {Shridhar, Mohit and Yuan, Xingdi and Côté, Marc-Alexandre and Bisk, Yonatan and Trischler, Adam and Hausknecht, Matthew},
  journal = {arXiv preprint arXiv:2010.03768},
  year = {2020},
  eprint = {2010.03768},
  archivePrefix = {arXiv},
  url = {https://arxiv.org/abs/2010.03768}
}

@article{eog,
  title = {{EnterpriseOps-Gym: Environments and Evaluations for Stateful Agentic Planning and Tool Use in Enterprise Settings}},
  author = {Malay, Shiva Krishna Reddy and Nayak, Shravan and Nair, Jishnu Sethumadhavan and Davasam, Sagar and Tiwari, Aman and Madhusudhan, Sathwik Tejaswi and Nemala, Sridhar Krishna and Sunkara, Srinivas and Rajeswar, Sai},
  journal = {arXiv preprint arXiv:2603.13594},
  year = {2026},
  eprint = {2603.13594},
  archivePrefix = {arXiv},
  url = {https://arxiv.org/abs/2603.13594}
}

@article{agentpoison,
  title = {{AgentPoison: Red-teaming LLM Agents via Poisoning Memory or Knowledge Bases}},
  author = {Chen, Zhaorun and Xiang, Zhen and Xiao, Chaowei and Song, Dawn and Li, Bo},
  journal = {arXiv preprint arXiv:2407.12784},
  year = {2024},
  eprint = {2407.12784},
  archivePrefix = {arXiv},
  url = {https://arxiv.org/abs/2407.12784}
}

@article{minja,
  title = {{Memory Injection Attacks on LLM Agents via Query-Only Interaction}},
  author = {Dong, Shen and Xu, Shaochen and He, Pengfei and Li, Yige and Tang, Jiliang and Liu, Tianming and Liu, Hui and Xiang, Zhen},
  journal = {arXiv preprint arXiv:2503.03704},
  year = {2025},
  eprint = {2503.03704},
  archivePrefix = {arXiv},
  url = {https://arxiv.org/abs/2503.03704}
}

@article{skillinject,
  title = {{Skill-Inject: Measuring Agent Vulnerability to Skill File Attacks}},
  author = {Schmotz, David and Beurer-Kellner, Luca and Abdelnabi, Sahar and Andriushchenko, Maksym},
  journal = {arXiv preprint arXiv:2602.20156},
  year = {2026},
  eprint = {2602.20156},
  archivePrefix = {arXiv},
  url = {https://arxiv.org/abs/2602.20156}
}

@article{maliciousskills,
  title = {{``Do Not Mention This to the User'': Detecting and Understanding Malicious Agent Skills in the Wild}},
  author = {Liu, Yi and Chen, Zhihao and Zhang, Yanjun and Deng, Gelei and Li, Yuekang and Ning, Jianting and Zhang, Leo Yu},
  journal = {arXiv preprint arXiv:2602.06547},
  year = {2026},
  eprint = {2602.06547},
  archivePrefix = {arXiv},
  url = {https://arxiv.org/abs/2602.06547}
}

@misc{agentskills,
  title = {Agent Skills Specification},
  author = {{Agent Skills}},
  year = {2026},
  howpublished = {\url{https://agentskills.io/specification}},
  note = {Pinned to commit 217be548739f21d6008915c29aefe320ea1a90af of https://github.com/agentskills/agentskills.}
}

@misc{aaif,
  title = {Donating {Agent Skills} to the {Agentic AI Foundation}},
  author = {{Anthropic}},
  year = {2026},
  howpublished = {\url{https://www.anthropic.com/news/donating-agent-skills-to-the-agentic-ai-foundation}}
}

@misc{mcp,
  title = {{Model Context Protocol} Specification, Revision 2026-07-28},
  author = {{Model Context Protocol}},
  year = {2026},
  howpublished = {\url{https://modelcontextprotocol.io/specification/2026-07-28}}
}

@misc{sep2640,
  title = {{SEP-2640} (draft): the io.modelcontextprotocol/skills extension},
  author = {{Model Context Protocol}},
  year = {2026},
  howpublished = {\url{https://github.com/modelcontextprotocol/modelcontextprotocol/commit/a3e147ca2710f68214247aecc729731ee1ae8d03}},
  note = {Draft specification enhancement proposal, pinned to the linked commit.}
}

@misc{claudeskills,
  title = {{Agent Skills} overview},
  author = {{Anthropic}},
  year = {2026},
  howpublished = {\url{https://platform.claude.com/docs/en/agents-and-tools/agent-skills/overview}},
  note = {Claude documentation.}
}

@misc{cowork,
  title = {{Claude Cowork} overview},
  author = {{Anthropic}},
  year = {2026},
  howpublished = {\url{https://claude.com/docs/cowork/overview}},
  note = {Claude documentation.}
}

@misc{codexskills,
  title = {{OpenAI Codex}: {MCP} skill provider ({OrchestratorSkillProvider})},
  author = {{OpenAI}},
  year = {2026},
  howpublished = {\url{https://github.com/openai/codex/blob/78c290807ce710180111df227df3b7a4fe845452/codex-rs/ext/skills/src/provider/orchestrator.rs}},
  note = {Source code, pinned to the linked commit.}
}

@misc{modelspec,
  title = {{OpenAI} Model Spec (2025-04-11): Ignore untrusted data by default},
  author = {{OpenAI}},
  year = {2025},
  howpublished = {\url{https://model-spec.openai.com/2025-04-11.html#ignore-untrusted-data-by-default}}
}

@misc{injection,
  title = {Mitigate jailbreaks and prompt injections},
  author = {{Anthropic}},
  year = {2026},
  howpublished = {\url{https://platform.claude.com/docs/en/test-and-evaluate/strengthen-guardrails/mitigate-jailbreaks}},
  note = {Claude documentation.}
}

\clearpage
\appendix
\needspace{10\baselineskip}
\section{Deviations from the original benchmarks}\label{deviations-from-the-original-benchmarks}

Table A1 lists each place where the experiments deviate from the original method, dataset, or harness of a benchmark.

\par\noindent\begin{minipage}{\linewidth}
\needspace{8\baselineskip}
\textbf{Table A1. Deviations from the original benchmarks.}\par\smallskip\noindent
\begin{tabular}{@{}
  >{\raggedright\arraybackslash}p{\dimexpr 0.1987\linewidth - 1.1922\tabcolsep\relax}
  >{\raggedright\arraybackslash}p{\dimexpr 0.2671\linewidth - 1.6026\tabcolsep\relax}
  >{\raggedright\arraybackslash}p{\dimexpr 0.2671\linewidth - 1.6026\tabcolsep\relax}
  >{\raggedright\arraybackslash}p{\dimexpr 0.2671\linewidth - 1.6026\tabcolsep\relax}@{}}
\toprule\noalign{}
\begin{minipage}[b]{\linewidth}\raggedright
Benchmark
\end{minipage} & \begin{minipage}[b]{\linewidth}\raggedright
Original
\end{minipage} & \begin{minipage}[b]{\linewidth}\raggedright
Deviation
\end{minipage} & \begin{minipage}[b]{\linewidth}\raggedright
Reason
\end{minipage} \\
\midrule\noalign{}
EnterpriseOps-Gym & The public dataset has 649 tasks, and the published leaderboard scores a task set that is not public. & The experiments use 625 tasks. & Tasks with known dataset defects are excluded. \\
EnterpriseOps-Gym & The ITSM policy makes \texttt{opened\_by} mandatory for a new incident. The ITSM server has no \texttt{opened\_by} column and no \texttt{opened\_by} argument. & A patch adds the field to the server. The policy text does not change. & An agent that obeys the policy cannot create the incident. \\
EnterpriseOps-Gym & The original checks. & The corrected task data removes the checks that an agent cannot pass if it obeys the task and the policy. The results give the original score and the corrected score. & Some checks need values that the task and the policy do not give, or check fields that the tool schema does not document. \\
EnterpriseOps-Gym & The public Stirrup harness converts the Gym tool schemas and sends \texttt{null} for unset optional arguments. & The harness sends the Gym schemas and the model arguments unchanged. For Anthropic models only, the harness removes a top-level \texttt{oneOf} from a tool schema. & Stirrup cannot convert the schemas of 27 tools (105 of 649 tasks). The calendar server returns HTTP 500 for a \texttt{null} optional argument. Anthropic rejects a top-level \texttt{oneOf}. \\
EnterpriseOps-Gym & The Stirrup default turn limit is 30. & The limit is 100. & The published leaderboard uses 100. \\
AppWorld & The benchmark reports results on its test splits. & The experiments use the Train and Dev scenarios. & The replay design needs teacher Traces of the scored tasks. \\
\ensuremath{\tau}\ensuremath{^3} & The upstream harness uses the native scaffold of each provider and strict text decoding of shell output. & The harness uses one agent loop, based on the OpenAI Responses API, for all executors and decodes shell output with replacement characters. & One scaffold for all models. Invalid UTF-8 in shell output stopped episodes. \\
ALFWorld & The RSD paper uses a fixed one-shot demonstration that it does not publish. & The experiments report two demonstrations, \texttt{fixed} and \texttt{typed}. & The demonstration changes the base score (Section 6.6). \\
ALFWorld & The RSD authors did not release code. The RSD paper does not give its ALFWorld version or its decoding settings. & We rebuilt the harness from the text of the RSD paper. & No code or settings were available. \\
ALFWorld & The ReAct demonstrations use \texttt{put\ X\ in/on\ Y}. & The demonstrations use \texttt{move\ X\ to\ Y}, and the prompt shows all admissible commands. & ALFWorld 0.4.2 accepts only \texttt{move\ X\ to\ Y}. \\
\bottomrule\noalign{}
\end{tabular}
\end{minipage}\par\medskip

\end{document}